\documentclass{article}

\usepackage{microtype}
\usepackage{graphicx}
\usepackage{subcaption}
\usepackage{booktabs} %

\usepackage{hyperref}

\usepackage[preprint]{icml2026}

\usepackage{amsmath}
\usepackage{amssymb}
\usepackage{mathtools}
\usepackage{amsthm}

\usepackage{amsmath,amsfonts,bm}

\def\eqref#1{(\ref{#1})}

\def\1{\bm{1}}

\def\vmu{{\bm{\mu}}}

\def\ve{{\bm{e}}}

\def\vm{{\bm{m}}}

\def\vs{{\bm{s}}}

\def\vv{{\bm{v}}}
\def\vw{{\bm{w}}}
\def\vx{{\bm{x}}}
\def\vy{{\bm{y}}}
\def\vz{{\bm{z}}}

\def\mA{{\bm{A}}}
\def\mB{{\bm{B}}}
\def\mC{{\bm{C}}}
\def\mD{{\bm{D}}}
\def\mE{{\bm{E}}}
\def\mF{{\bm{F}}}

\def\mI{{\bm{I}}}

\def\mK{{\bm{K}}}

\def\mM{{\bm{M}}}

\def\mQ{{\bm{Q}}}

\def\mS{{\bm{S}}}

\def\mV{{\bm{V}}}

\def\mX{{\bm{X}}}

\def\mZ{{\bm{Z}}}

\def\mSigma{{\bm{\Sigma}}}

\DeclareMathAlphabet{\mathsfit}{\encodingdefault}{\sfdefault}{m}{sl}
\SetMathAlphabet{\mathsfit}{bold}{\encodingdefault}{\sfdefault}{bx}{n}

\def\gN{{\mathcal{N}}}

\newcommand{\E}{\mathbb{E}}

\newcommand{\R}{\mathbb{R}}

\DeclareMathOperator*{\argmin}{arg\,min}

\DeclareMathOperator{\Tr}{Tr}

\usepackage{float}

\usepackage{enumitem}

\usepackage{caption}
\usepackage{subcaption}
\usepackage{csquotes}
\MakeOuterQuote{"}

\usepackage{tabularx}
\usepackage{booktabs}

\usepackage{xcolor}

\usepackage[capitalize,noabbrev]{cleveref}

\theoremstyle{plain}
\newtheorem{theorem}{Theorem}[section]
\newtheorem{proposition}[theorem]{Proposition}
\newtheorem{lemma}[theorem]{Lemma}
\newtheorem{corollary}[theorem]{Corollary}
\theoremstyle{definition}
\newtheorem{definition}[theorem]{Definition}
\newtheorem{assumption}[theorem]{Assumption}
\theoremstyle{remark}
\newtheorem{remark}[theorem]{Remark}

\usepackage[textsize=tiny]{todonotes}

\icmltitlerunning{	
Optimal Attention Temperature Improves the Robustness of In-Context Learning under Distribution Shift in High Dimensions}

\begin{document}

\twocolumn[
  \icmltitle{	
Optimal Attention Temperature Improves the Robustness\\ of In-Context Learning under Distribution Shift in High Dimensions}

  \icmlsetsymbol{equal}{*}

  \begin{icmlauthorlist}
    \icmlauthor{Samet Demir}{mlipkuis}
    \icmlauthor{Zafer Doğan}{mlipkuis,ee}
  \end{icmlauthorlist}

  \icmlaffiliation{mlipkuis}{MLIP Research Group, KUIS AI Center, Koç University}
  \icmlaffiliation{ee}{Department of EEE, Koç University, İstanbul, Turkey}

  \icmlcorrespondingauthor{Zafer Doğan}{zdogan@ku.edu.tr}

  \icmlkeywords{Machine Learning, ICML}

  \vskip 0.3in
]

\printAffiliationsAndNotice{}  %

\begin{abstract}
Pretrained Transformers can perform in-context learning (ICL) from a few demonstrations, but this ability can fail sharply when the test distribution differs from pretraining—a common deployment setting. We study attention temperature as a simple inference-time control for improving ICL robustness under such shifts. In a high-dimensional linear-regression framework, we analyze a Transformer with "approximate softmax" attention, which preserves softmax's normalization and temperature-dependent selectivity while remaining tractable. We derive a closed-form expression for the ICL generalization error under distribution shift, and show that it is minimized by an explicit optimal attention temperature. This characterization yields interpretable guidance by linking the best temperature to moments of the pre-softmax attention scores, and predicts when temperature adjustment can recover near Bayes-optimal performance. We validate the theory with extensive simulations, and further demonstrate gains on pretrained LLMs (GPT-2 and Llama2-7B) on question-answering benchmarks under distribution shift induced by noisy in-context demonstrations. Overall, attention temperature emerges as a principled, lightweight knob for improving the robustness of ICL in pretrained Transformers.
\end{abstract}

\section{Introduction}

Transformers \citep{vaswani2017attention} have emerged as the foundational architecture of contemporary AI systems, underpinning state-of-the-art models such as ChatGPT, Gemini, and DeepSeek. Central to their remarkable success is \emph{in-context learning} (ICL)—the capability to adapt to novel tasks directly from prompts without any gradient-based weight updates \citep{brown2020language}. This property, often described as emergent, has catalyzed a surge of research aimed at uncovering the underlying mechanisms of ICL \citep{akyurek2023what, pmlr-v202-von-oswald23a} and at characterizing how factors such as task diversity and model scale shape its performance \citep{wei2022emergent, wu2024how}.

Yet, despite its transformative potential, ICL exhibits pronounced sensitivity to distributional shifts between pretraining and downstream tasks. Both empirical and theoretical investigations demonstrate that even mild shifts can substantially degrade performance \citep{zhang2024trained}, underscoring unresolved questions about the robustness, generalization, and adaptability of pretrained Transformer models. Addressing these limitations is crucial for realizing the full promise of ICL in reliable, deployable AI systems.

At the core of the Transformer architecture is the self-attention mechanism, defined as  
\begin{align}
    \text{Attention}(\mZ) 
    := 
    \mV \mZ \cdot 
    \text{softmax}\!\left( 
        \frac{(\mK \mZ)^T (\mQ \mZ)}{\tau} 
    \right),
    \label{eq:softmax_attention}
\end{align}
where $\mZ$ denotes the input representation and $\mQ$, $\mK$, and $\mV$ are the query, key, and value weight matrices, respectively.  
The parameter $\tau>0$, referred to as the \emph{attention temperature}, controls the variance of the softmax outputs and hence the selectivity of attention weights.  
This quantity is distinct from the “sampling temperature” commonly used to adjust the output distribution of generative models such as large language models (LLMs) \citep{renze2024effect}.  
Throughout this work, we exclusively focus on the attention temperature as an intrinsic component of the attention mechanism.  

While the original Transformer fixes $\tau=\sqrt{d_k}$ \citep{vaswani2017attention}, where $d_k$ is the key dimension, subsequent empirical studies have demonstrated that adjusting the attention temperature can enhance performance across diverse NLP and computer vision benchmarks \citep{lin2018learning, zhang2022attention, peng2024yarn, lee2021vision, chen2023accumulated, zou2024attention}.  
Yet, to the best of our knowledge, its role within \emph{in-context learning} (ICL) remains unexplored.  
Because the attention temperature directly governs how sharply the model concentrates on specific inputs, it is poised to critically influence ICL behavior under distributional shift—a setting of central practical relevance, where mismatches between training and inference distributions are ubiquitous.

\vspace{-1em}
\paragraph{This work ---}  
In this paper, we present a unified theoretical and empirical study of the \emph{attention temperature} in the context of in-context learning (ICL).  
We focus on how adjusting this parameter can systematically improve the ICL performance of pretrained Transformers under distributional shift.  
We address this question in the setting of linear regression tasks, which offer a well-controlled yet expressive framework for dissecting the mechanisms of ICL \citep{garg2022can, zhang2024trained}.  
Departing from prior work restricted to linear attention, we analyze a Transformer with \emph{approximate softmax} attention—an architecture that preserves the essential temperature-dependent behavior of standard attention while remaining mathematically tractable.

Our analysis yields a closed-form characterization of the \emph{optimal temperature}—the value of $\tau$ that minimizes generalization error during inference.  
We show that this optimal temperature depends explicitly on the nature of the distribution shift and that setting it appropriately can recover or even surpass baseline ICL performance.  
We validate our theoretical predictions through extensive experiments on both synthetic (linear regression) and real-world (question answering with LLMs) tasks, demonstrating that temperature selection constitutes a simple yet powerful mechanism for improving robustness.

\vspace{-1em}
\paragraph{Contributions ---}  
Our work makes the following key contributions:
\begin{enumerate}[topsep=0pt,itemsep=-1ex,partopsep=1ex,parsep=1ex,leftmargin=0.6cm]
    \item We provide, to our knowledge, the first theoretical characterization of the optimal attention temperature for pretrained Transformers with \emph{approximate softmax attention} in ICL tasks.
    \item We analyze the generalization behavior of such models under a broad spectrum of distributional shifts, employing weaker assumptions than prior studies.
    \item We establish a clear theoretical and empirical link between distribution shift and attention temperature, showing that principled temperature selection can substantially enhance ICL performance across diverse tasks.
\end{enumerate}

Taken together, these results offer new insights into the interplay between temperature, distribution shift, and generalization in in-context learning, and highlight a practical avenue for improving the robustness of pretrained Transformers.
\vspace{-1em}
\paragraph{Notation ---} We follow standard notation from \citet{goodfellow2016deep}. The spectral norm of matrix $\mM$ is denoted by $\|\mM\|$, and the trace by $\text{Tr}(\mM)$. Matrix entries and slices are denoted as \( M_{i,j} \), \( \mM_{:,j} \), and \( \mM_{i,:} \).

\section{Related work}
\paragraph{Theory of in-context learning ---} 
Simplified Transformer variants—particularly those using linear attention—have proven useful for gaining analytical insights about ICL \citep{garg2022can, zhang2024trained, raventos2024pretraining}. Notably, \citet{zhang2024trained} showed that linear Transformers approximate Bayes-optimal inference in linear regression tasks, even under distribution shift. We build on this line of research but focus explicitly on the role of the \emph{attention temperature}.  
In contrast to \citet{zhang2024trained}, we (i) employ \emph{approximate softmax attention} to isolate the effect of temperature, (ii) study how temperature adjustments can mitigate the impact of distribution shifts, and (iii) derive and empirically evaluate the \emph{optimal temperature} for improving ICL performance.  
These advances extend prior analyses and yield a deeper theoretical and empirical understanding of how principled temperature selection enhances the robustness of Transformers under distributional shift.\footnote{Due to space limitations, additional related work is discussed in Appendix~\ref{appendix:related_work}.}

\paragraph{Linear vs. softmax attention ---}
Although linear attention has gained traction for its computational efficiency, it typically lags behind softmax-based counterparts in predictive performance, spurring efforts to narrow this gap \citep{choromanski2021rethinking, zhen2022cosformer}.  
A pivotal advance in this direction is due to \citet{han2024bridging}, who showed that a \emph{approximate variant of softmax attention} can closely approximate the performance of standard softmax attention.  
Building on this insight, we adopt the \emph{approximate softmax} formulation, which preserves the essential temperature-dependent behavior of standard attention while enabling tractable theoretical analysis.  
This choice provides a principled framework for investigating how attention-temperature selection shapes ICL performance in pretrained Transformers.
\vspace{-1em}
\paragraph{Attention temperature ---}
Research on attention temperature remains limited.  
\citet{velivckovic2024softmax} recently proposed an adaptive temperature scheme to sharpen softmax outputs, and several empirical studies in natural language processing and computer vision \citep{lin2018learning, zhang2022attention, peng2024yarn, lee2021vision, chen2023accumulated, zou2024attention} suggest that adjusting the attention temperature can enhance Transformer performance.  
However, these works do not examine ICL under distributional shift.  
To our knowledge, no prior study has systematically analyzed how attention temperature influences ICL in such settings—a gap our work directly addresses.

\begin{figure*}[h]
    \centering
    \includegraphics[width=0.99\linewidth]{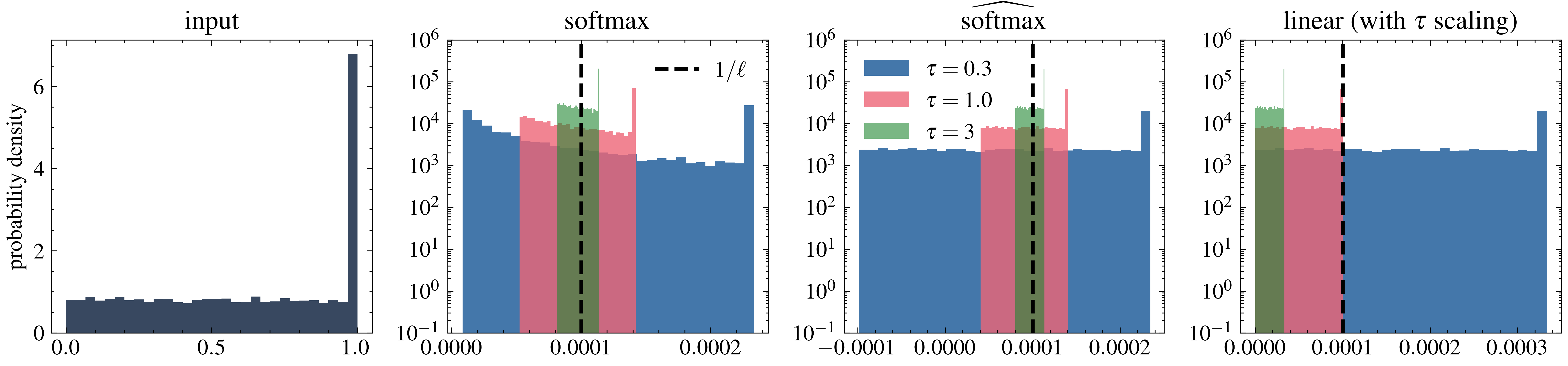}
    \caption{Comparison of temperature effects in softmax, approximate softmax, and linear (with temperature scaling) cases. We consider an input vector $\vx \in \R^l$ whose histogram is illustrated on the left-most plot. Rest of the plots illustrates histograms of the elements of $\text{softmax}(\vx / \tau)$, approximation $\widehat{\text{softmax}}(\vx /\tau)$ derived in Appendix \ref{appendix:approximate_softmax} and $\vx/(l \tau)$ from left to right, respectively.}
    \label{fig:temperature_effects_approximate_vs_softmax}
\end{figure*}

\vspace{-1em}
\section{Setting}

We describe the setup for analyzing ICL in linear regression using pretrained Transformers, covering the data model, approximate softmax attention with reparameterization, evaluation metrics, and the Bayes-optimal benchmark.

\subsection{In-context learning for linear regression}
We study the ICL abilities of pretrained Transformers on linear regression tasks. Given a sequence of tokens, i.e.,  input-label pairs,
\( \{(\vx_1, y_1), (\vx_2, y_2), \dots, (\vx_{l-1}, y_{l-1}), (\vx_{l},?)\}  \)
where each input vector \( \vx_i \in \mathbb{R}^d \) and corresponding label \( y_i \in \mathbb{R} \) are independently sampled from an unknown joint distribution, the model must predict $y_l$ using only the context $\{(\vx_i, y_i)\}_{i=1}^{l-1}$ and the query $\vx_l$, where \( l-1 \) is referred as the "context length". Each $(\vx_i, y_i)$ pair is sampled i.i.d. from a joint distribution defined by:
\begin{align}
    \vx_i \sim \gN(\vmu_x, \mSigma_x), \;
    y_i = \vw^T \vx_i + \epsilon_i, \;
    \epsilon_i \sim \gN(0, \sigma^2),
    \label{eq:data_model}
\end{align}
where the task vector $\vw \sim \gN(\vmu_w, \mSigma_w)$ is fixed within a context but varies across tasks.

\begin{assumption}[Well-behaved data distributions]
\label{assumption:mean_cov_bound}
There exist constants $c_1, c_2, c_3 > 0$ such that:
\begin{align}
    &\|\vmu_x\|, \|\vmu_w\| \leq c_1, \quad
\lambda_{\min}(\mSigma_x), \lambda_{\min}(\mSigma_w) \geq c_2,\\
&\lambda_{\max}(\mSigma_x), \lambda_{\max}(\mSigma_w) \leq c_3.
\end{align}
\end{assumption}
This assumption ensures that the input and task distributions have bounded means and covariances, offering greater flexibility than the more restrictive setup of \citet{zhang2024trained}.
\begin{assumption}[High-dimensional regime]
\label{assumption:jointly_diverge}
The context length $l$ and input dimension $d$ diverge jointly: $l, d \to \infty$.
\end{assumption}
This assumption reflects realistic settings where both context length and input dimension grow simultaneously, aligning with modern ML trends and enabling analysis of generalization in high-dimensional regimes.

Under this set of assumptions, we define ICL for linear regression tasks as follows: 
\begin{definition}[In-Context Learning (ICL)]
\label{definition:icl}
A model succeeds at ICL for linear regression if its generalization error nearly matches that of the Bayes-optimal linear model (defined in Section \ref{section:bayes_optimal}).
\end{definition}

\subsection{Modeling attention with transformers}
Following the convention established by \citet{zhang2024trained}, we represent the input sequence by an embedding matrix:
\begin{align}
    \mZ := \begin{bmatrix}
        \vx_1 & \cdots & \vx_{l-1} & \vx_l \\
        y_1 & \cdots & y_{l-1} & 0
    \end{bmatrix} \in \mathbb{R}^{(d+1) \times l},
    \label{eq:embedding_matrix}
\end{align}
where the last column corresponds to the query input with no label.

Given this embedding, the softmax self-attention output is
\begin{align}
    \mS := \mZ + \mV \mZ \cdot \text{softmax}\left( \frac{(\mK \mZ)^T (\mQ \mZ)}{\tau} \right),
    \label{eq:softmax_attention_with_residual}
\end{align}
where $\mK$, $\mQ$, and $\mV$ are the key, query, and value matrices, respectively, and $\tau$ is the temperature.

Here, we denote the model’s prediction as $S_{d+1, l}$ — the last element in the final row.

\subsection{Approximate softmax attention}
To analytically characterize the effect of temperature on ICL, we employ an approximation of softmax attention :
\begin{align}
    \mE := \mZ + \mV \mZ \cdot \widehat{\text{softmax}}\left( \frac{(\mK \mZ)^T (\mQ \mZ)}{\tau} \right),
    \label{eq:approximate_attention}
\end{align}
the explicit form of which is provided together with its derivation in Appendix \ref{appendix:approximate_softmax}. In contrast to linear attention \citep{zhang2024trained},
\begin{align}
    \mZ + \frac{1}{l} \mV \mZ (\mK \mZ)^T (\mQ \mZ),
    \label{eq:linear_attention}
\end{align}
our formulation in \eqref{eq:approximate_attention} explicitly preserves normalization, which is essential for both interpretability and robustness. This difference is described in the following remark.

\begin{remark}[Linear vs. approximate softmax attention]
\label{remark:linear_vs_approximate}
Approximate softmax attention maintains row-wise normalization, making it inherently more robust to shifts in input means — a critical failure mode of linear attention in ICL.
\end{remark}

Another key distinction between the linear case and the approximate softmax case is that linear (with temperature scaling) fails to capture the temperature behavior of softmax. However, while this may not be immediately apparent, approximate softmax closely mirrors the behavior of softmax with respect to temperature variation. Figure \ref{fig:temperature_effects_approximate_vs_softmax} visualizes the effects of temperature across softmax, approximate softmax, and linear (with temperature scaling) settings under an example input. The figure demonstrates our claim that approximate softmax can be a useful tool to study the temperature effect exhibited by softmax. We expand on this point in Appendix~\ref{appendix:temperature_in_approximate_softmax}.

\subsection{Reparametrization of approximate softmax attention}

To streamline analysis, we reparametrize the matrices $\mV$ and $\mM := \mK^T \mQ$ as:
\begin{align}
    \mV = \begin{bmatrix}
        * & * \\
        \vv_{21}^T & v_{22}
    \end{bmatrix}, \quad
    \mM = \begin{bmatrix}
        \mM_{11} & * \\
        \vm_{21}^T & *
    \end{bmatrix},
    \label{eq:reparametrization}
\end{align}
where only $\vv_{21}$, $v_{22}$, $\vm_{21}$, and $\mM_{11}$ influence the prediction $\hat{y}(\mZ; \mV, \mM)$. The remaining terms are denoted by \( * \) as they are not relevant for predicting \( y_l \) in this context. The prediction from the Transformer model \eqref{eq:approximate_attention} can thus be expressed as a function of \(\mM\) and \(\mV\), i.e., \( \hat{y}(\mZ; \mV, \mM) := E_{d+1, l} \). This form parallels the approach by \citet{zhang2024trained}, allowing for tractable theoretical analysis.

By analyzing this reparameterization, we gain a deeper understanding of how the model parameters interact with the data to address the ICL problem effectively. %

\subsection{Evaluating generalization performance}
We focus on evaluating the performance of our attention model by assessing its generalization error, measuring the ICL performance. For a given set of parameters \((\mV, \mM)\), the model's generalization (ICL) error is defined as:
\begin{align}
    \mathcal{G}(\mV, \mM) := \underset{(\mZ, y_{l}) \sim \mathcal{D}^{test}}{\mathbb{E}} \left[ \left( y_{l} - \hat{y}(\mZ;\mV, \mM) \right)^2 \right],
    \label{eq:generalization_definition}
\end{align}
where \( \mathcal{D}^{test} \) denotes the distribution of the test set, which includes input-output pairs generated with tasks that the model has not encountered during training. Since the task vectors in the test set differ from those encountered during training, the model is required to infer these new vectors based solely on the provided context. Therefore, the ICL/generalization error \eqref{eq:generalization_definition} assesses the genuine ICL capabilities of the model.

\subsection{Bayes-optimal ridge estimator}
\label{section:bayes_optimal}
The Bayes-optimal ridge estimator provides a principled framework for estimating the task vector \(\vw\) given a prior distribution and a set of \(l -1 \) samples. It is defined as:
\begin{align}
    \hat{\vw}_{Bayes} =   \left(\frac{\bar{\mX}^T \bar{\mX}}{\sigma^2} + \mSigma_w^{-1}\right)^{-1} \left(\frac{\bar{\mX}^T \bar{\vy}}{\sigma^2} + \mSigma_w^{-1} \vmu_w\right),
    \label{eq:bayes_optimal_ridge_estimator}
\end{align}
where \(\bar{\mX}\) is the centered input matrix and \(\bar{\vy}\) is the centered label vector. This estimator combines information from observed data with prior knowledge of the distribution of \(\vw\), thereby balancing bias and variance. It serves as the gold standard against which we benchmark model predictions. The terms involving \(\mSigma_w^{-1}\) introduce a regularization effect, which is particularly advantageous in high-dimensional regimes.

The derivation, provided in Appendix~\ref{appendix:bayes_optimal}, illustrates how Bayesian principles inform regression by integrating data evidence with prior distributions to yield more reliable predictions. In our setting, the inputs and labels are derived from the prompt matrix \(\mZ\), and the Bayes-optimal linear model predicts any input \(\vx\) as \(\hat{\vw}_{Bayes}^T \vx\).

\begin{figure*}[t]
    \centering
    \begin{subfigure}[b]{0.32\textwidth}
         \centering
         \includegraphics[width=0.99\linewidth]{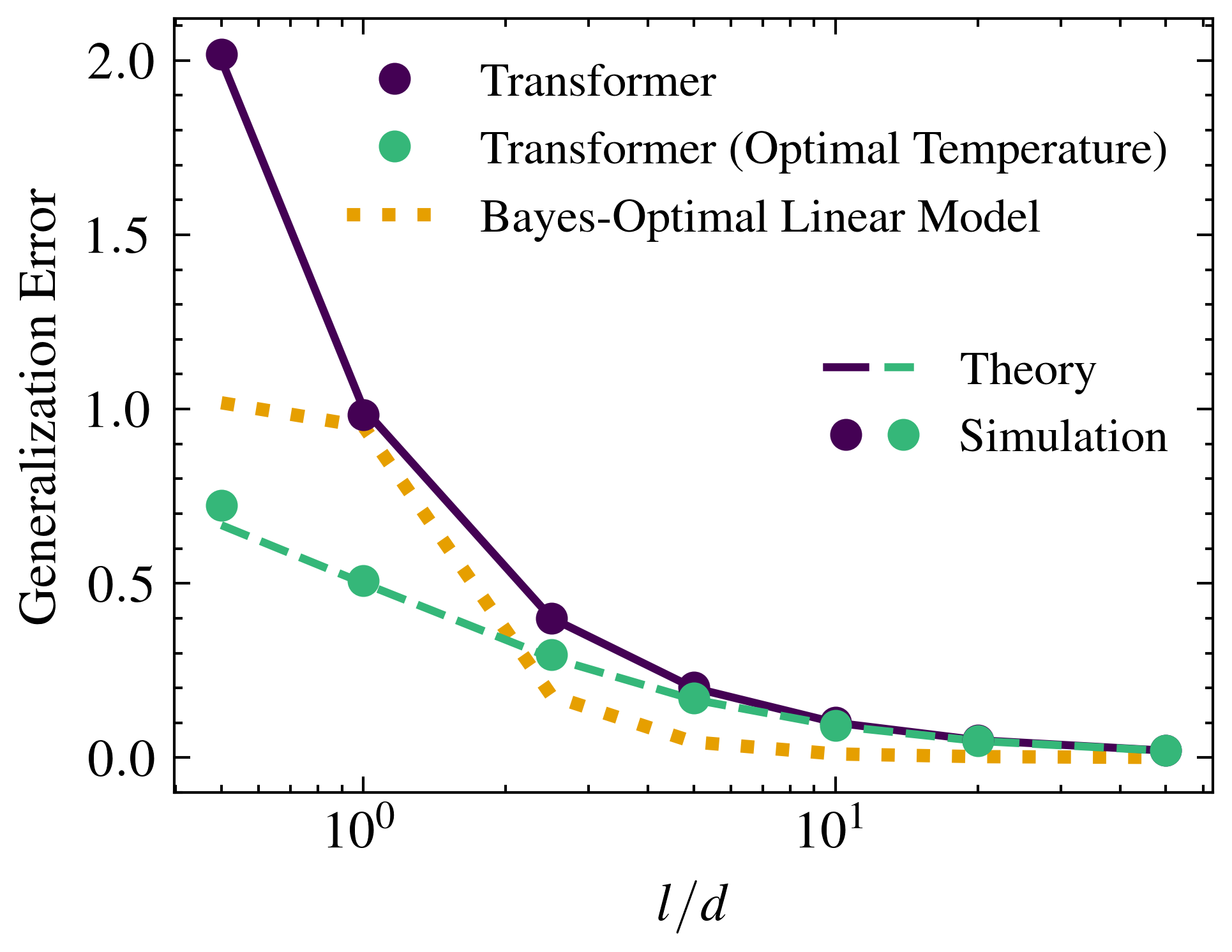}
         \captionsetup{justification=centering}
         \caption{Without\\Distribution Shift\\($\mathcal{D}^{train} = \mathcal{D}^{test}$)}
     \end{subfigure}
     \begin{subfigure}[b]{0.32\textwidth}
         \centering
         \includegraphics[width=0.99\linewidth]{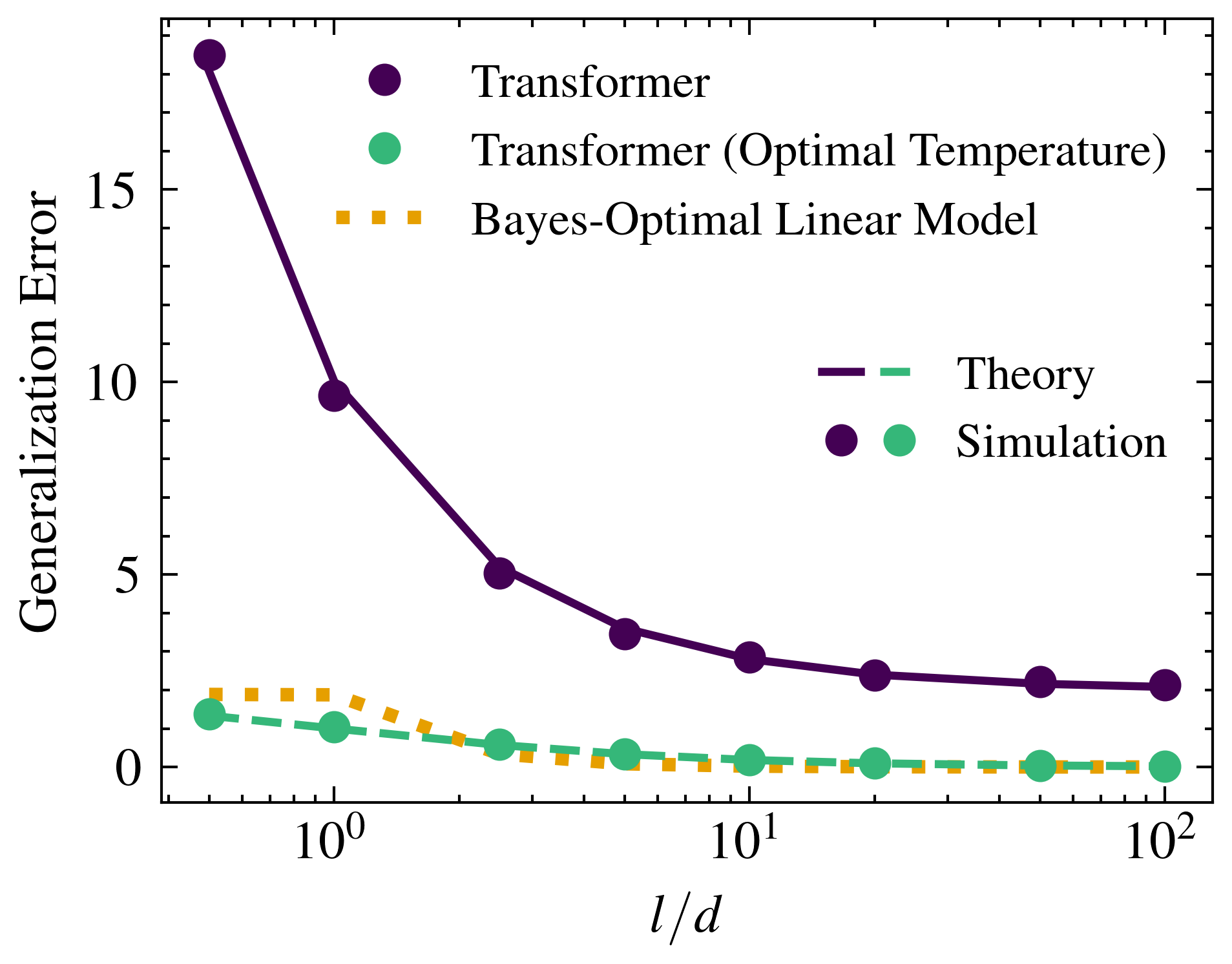}
         \captionsetup{justification=centering}
         \caption{With Shift in\\Input Covariance\\
         ($\mSigma_{test} = 2 \mSigma_{train}$)}
     \end{subfigure}
      \begin{subfigure}[b]{0.31\textwidth}
         \centering
         \includegraphics[width=0.99\linewidth]{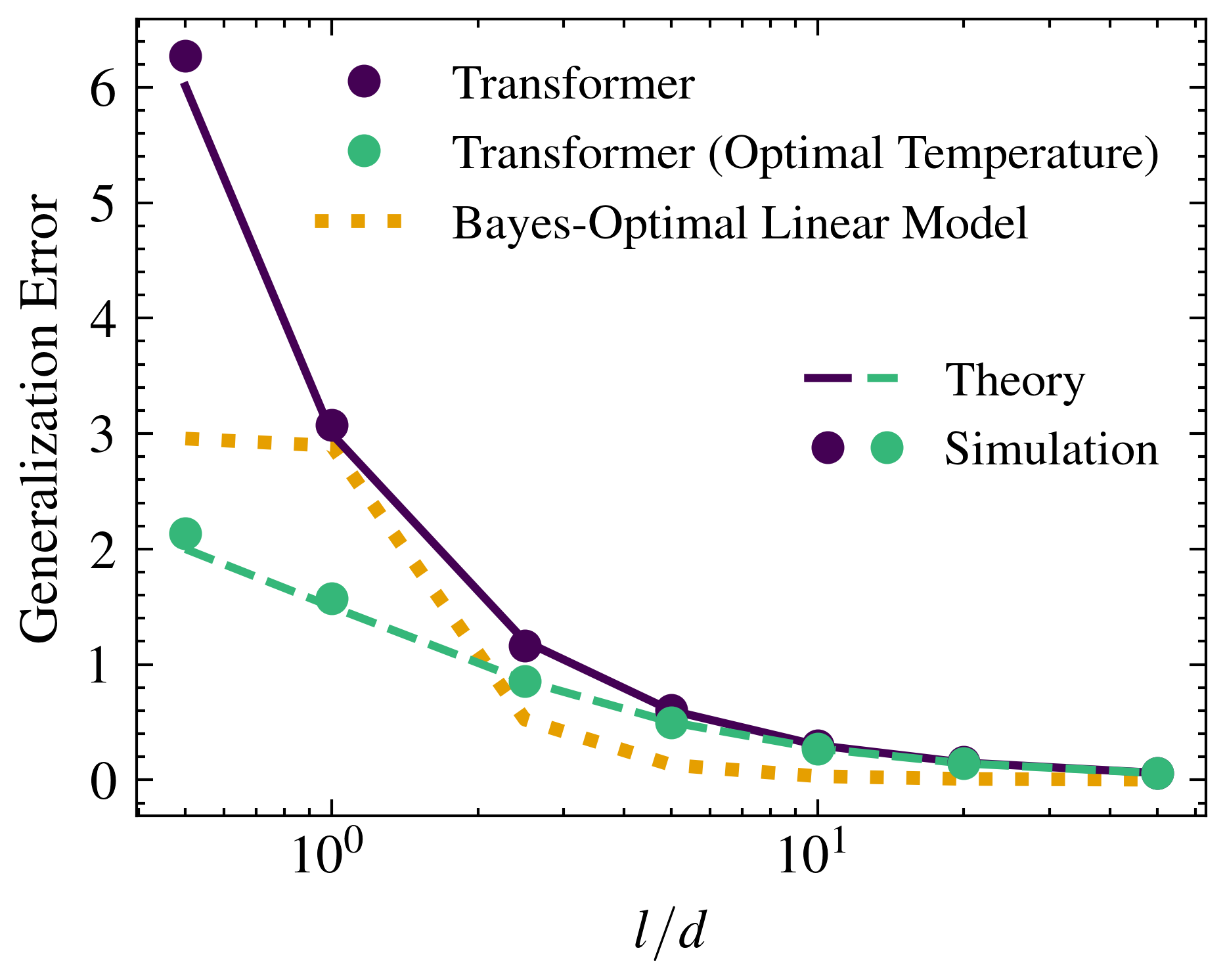}
         \captionsetup{justification=centering}
         \caption{With Shift in\\ Task Distribution\\
         ($\vmu_w^{test} = \mathbf{1}/\sqrt{d}$, $\mSigma_w^{test} = 3 \mSigma_w^{train}$)}
     \end{subfigure}
     \caption{Experiments with Transformer \eqref{eq:approximate_attention} on ICL under distribution shifts. Parameters are set using \eqref{eq:pretraining} while the optimal temperature is calculated by Theorem~\ref{theorem:optimal_temperature}. Here, $d=50$, $m=5000$ (with a new task per sample), $\sigma = 0.1$, $\vmu_x^{train} = \vmu_w^{train} = \mathbf{0}$, and $\mSigma_x^{train} = \mSigma_w^{train} = \mI$.}
     \label{fig:approximate_attention_experiments}
\end{figure*}

\section{Theoretical results}

In this section, we present our main theoretical results on the ICL under distribution shifts for the Transformer with an approximate softmax attention without MLP layers, denoted by \eqref{eq:approximate_attention}.  We provide an asymptotic characterization of its generalization error, offering a principled framework for analyzing performance. Next, we investigate the role of the temperature parameter and demonstrate that adjusting it appropriately can substantially improve generalization—especially in cases where the model initially fails to perform effective in-context learning. We then identify specific conditions for which the model fails to generalize under distribution shifts at test time, revealing key limitations of the model in ICL. We describe the interpretation of the found analytical form of the optimal attention temperature in mitigating the adverse effects of distribution shift. Finally, we explain a possible extension of our optimal temperature to other ICL settings involving softmax attention.

\subsection{In-context learning performance}
We analyze the in-context learning (ICL) performance of the Transformer model \eqref{eq:approximate_attention} by evaluating the generalization error defined in~\eqref{eq:generalization_definition}. To establish a general setting for the subsequent results, we impose the following assumption on the pretrained parameters.
\begin{assumption}\label{asssumption:parameters_M_V}
There exists a constant $c > 0$ such that
\[
\|\mM_{11}\| \leq c d, \quad \|\vm_{21}\| = 0, \quad \|\vv_{21}\| \leq \frac{c}{d l}, \quad |v_{22}| \leq \frac{c}{d}.
\]
\end{assumption}
The generalization error result stated below holds for any parameters $\mM, \mV$ that satisfy Assumption~\ref{asssumption:parameters_M_V}.

\begin{theorem}[Generalization error for ICL]
Suppose Assumptions~\ref{assumption:mean_cov_bound}--\ref{assumption:jointly_diverge} and~\ref{asssumption:parameters_M_V} hold. At test time, assume the input, task, and noise distributions are given by $\mathcal{N}(\vmu_x, \mSigma_x)$, $\mathcal{N}(\vmu_w, \mSigma_w)$, and $\mathcal{N}(0, \sigma^2)$, respectively. Define
\[
\mA := \mSigma_x + \vmu_x \vmu_x^T, \quad \mB := \mSigma_w + \vmu_w \vmu_w^T.
\]
Then, the generalization error is
\begin{align}
\mathcal{G}(\mV, \mM) &= \frac{1}{\tau^2} \Tr\left(\mA \mM_{11}^T \mF_1 \mM_{11} \right) \nonumber\\
&\quad- \frac{1}{\tau} \Tr\left(\mA \left( \mF_2 \mM_{11} + \mM_{11}^T \mF_2^T \right) \right) \nonumber\\
&\quad+ \Tr\left(\mA \mB \right) + \sigma^2,
\end{align}
where
\begin{align}
\mF_1 &:= \left( \mSigma_x \hat{\mB} + \frac{1}{l} \left( v_{22}^2 \sigma^2 + \Tr(\hat{\mB} \mSigma_x) \right) \mI \right) \mSigma_x,\\ 
\mF_2 &:= (\vmu_w \vv_{21}^T + v_{22} \mB) \mSigma_x,\\
\hat{\mB} &:= v_{22} \vmu_w \vv_{21}^T + v_{22} \vv_{21} \vmu_w^T + v_{22}^2 \mB.
\end{align}

\begin{proof}
The generalization error is derived using Isserlis’ theorem~\citep{isserlis1918} to compute higher-order moments. See Appendix~\ref{appendix:generalization_error} for the full derivation.
\end{proof}
\label{theorem:generalization_error}
\end{theorem}

Theorem~\ref{theorem:generalization_error} illustrates how the parameters $\mM$, $\mV$, and the test-time data distribution affect the generalization error. Notably, the temperature parameter $\tau$ plays a critical role.

Although temperature can be implicitly encoded in $\mM$ during pretraining, it becomes especially important under distribution shifts that the model is not equipped to handle. In such cases, one can optimize generalization performance by choosing the temperature $\tau_{\text{opt}}$ that minimizes the generalization error, as discussed next.

\subsection{Optimal attention temperature}
\label{section:optimal_temperature}

To address distribution shifts, we define the optimal attention temperature as follows:

\begin{theorem}[Optimal attention temperature]\label{theorem:optimal_temperature}
Suppose Assumptions \ref{assumption:mean_cov_bound}, \ref{assumption:jointly_diverge}, and \ref{asssumption:parameters_M_V} hold. To minimize the generalization error, the optimal attention temperature for inference is given by
\begin{align}
    \tau_{\text{opt}} = \frac{ 2\text{Tr}\left(\mA \mM_{11}^T \mF_1 \mM_{11}\right)}{\text{Tr}\left(\mA \left(\mF_2 \mM_{11} + \mM_{11}^T \mF_2^T\right)\right)},
    \label{eq:optimal_temperature}
\end{align}
provided that $\text{Tr}\left(\mA \left(\mF_2 \mM_{11} + \mM_{11}^T \mF_2^T\right)\right)>0$ and $\text{Tr}\left(\mA \mM_{11}^T \mF_1 \mM_{11}\right)>0$.
\begin{proof}
    We minimize the generalization error from Theorem \ref{theorem:generalization_error} with respect to $\tau$ (Appendix \ref{appendix:optimal_temperature}).
\end{proof}
\end{theorem}

Consider the optimal temperature $\tau_{\text{opt}}$ from Theorem \ref{theorem:optimal_temperature}. When $\tau_{\text{opt}} \neq 1$, using an unadjusted temperature leads to suboptimal generalization error. Thus, incorporating the optimal temperature improves generalization in in-context learning under distribution shift.

\begin{figure*}[t]
    \centering
    \begin{subfigure}[b]{0.32\textwidth}
         \centering
         \includegraphics[width=0.99\linewidth]{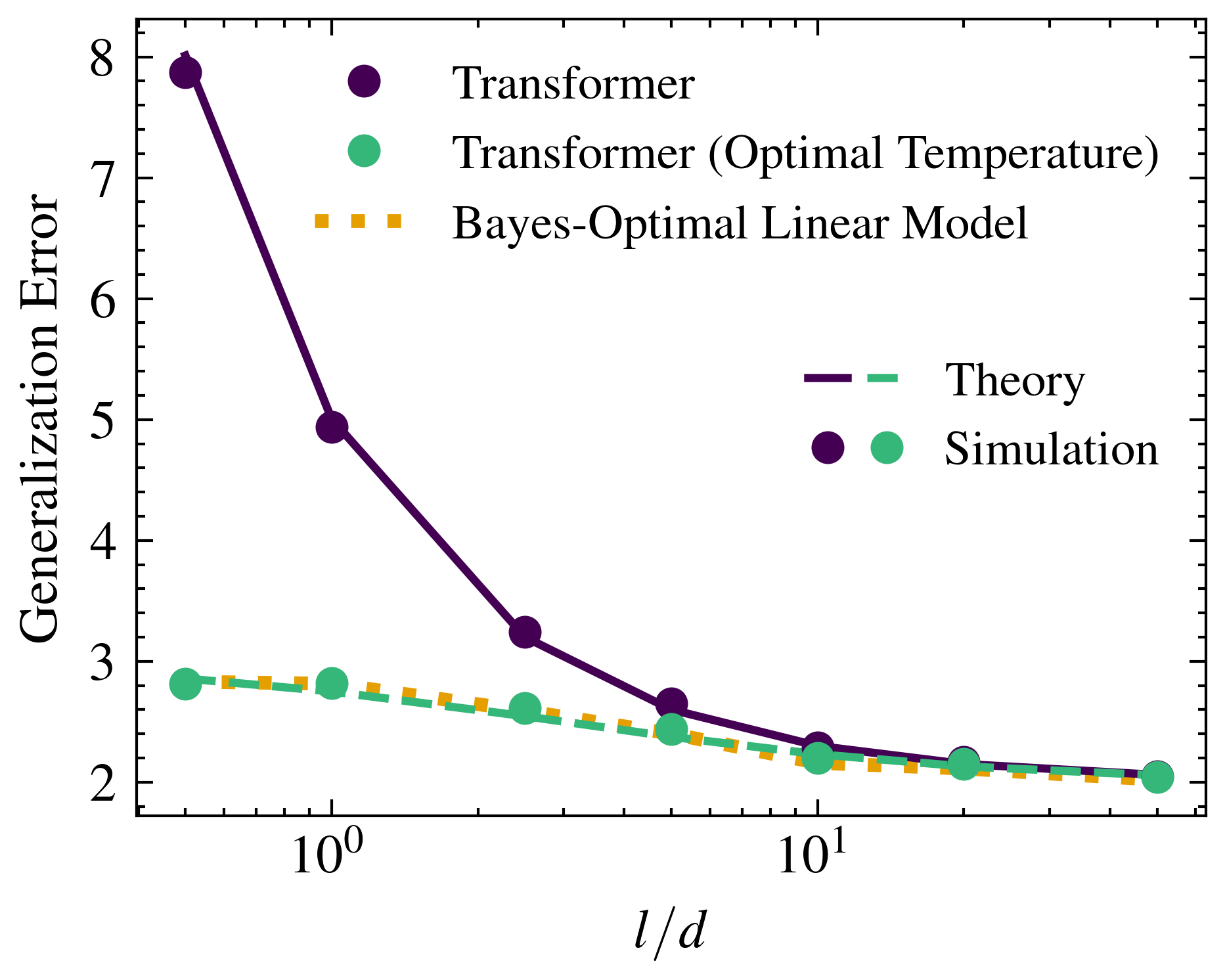}
         \captionsetup{justification=centering}
         \caption{Effect of $l/d$ when $\sigma_{test} = 10$}
     \end{subfigure}
     \begin{subfigure}[b]{0.32\textwidth}
         \centering
         \includegraphics[width=0.99\linewidth]{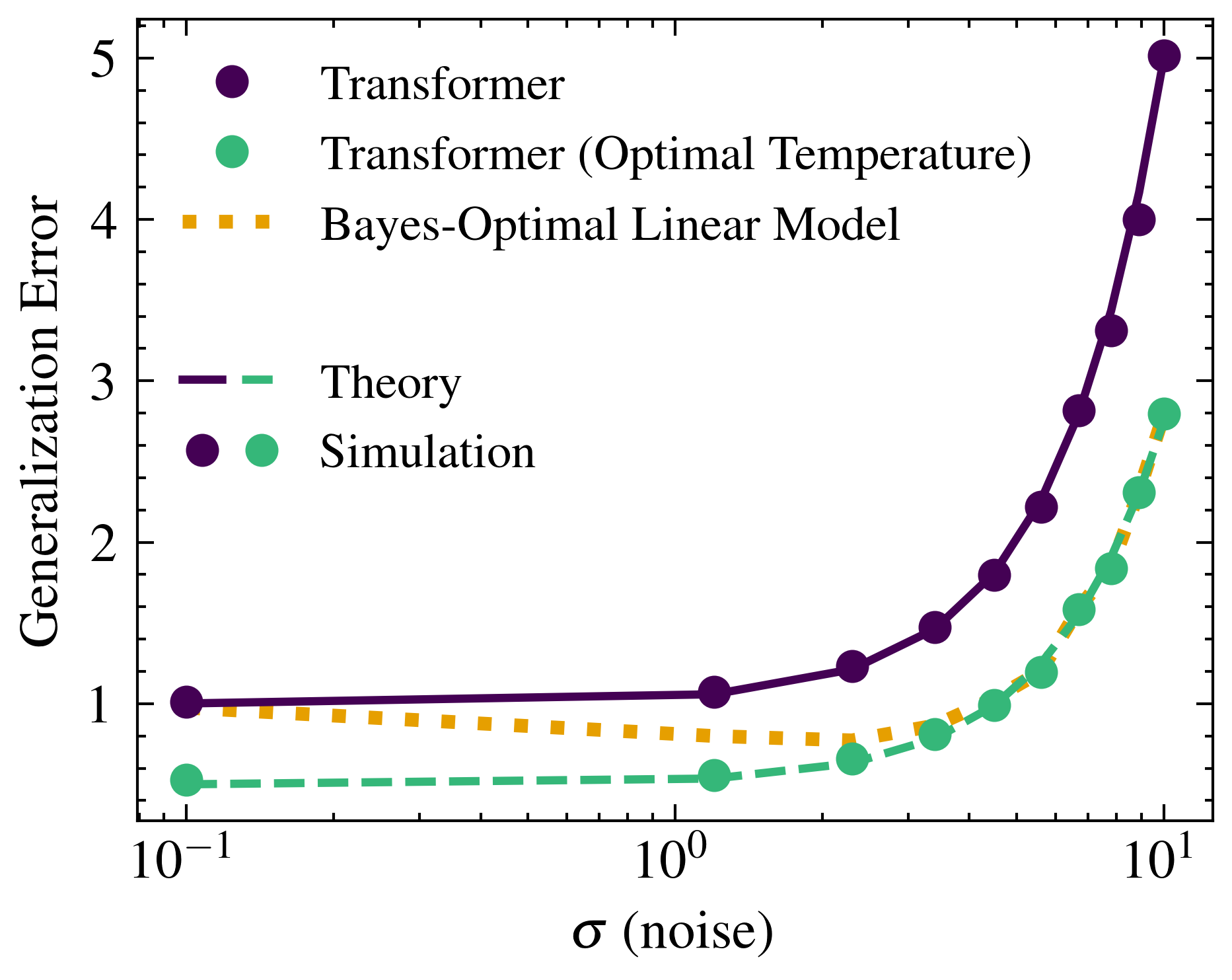}
         \captionsetup{justification=centering}
         \caption{Effect of $\sigma_{test}$ when $l/d = 1$}
     \end{subfigure}
      \begin{subfigure}[b]{0.329\textwidth}
         \centering
         \includegraphics[width=0.99\linewidth]{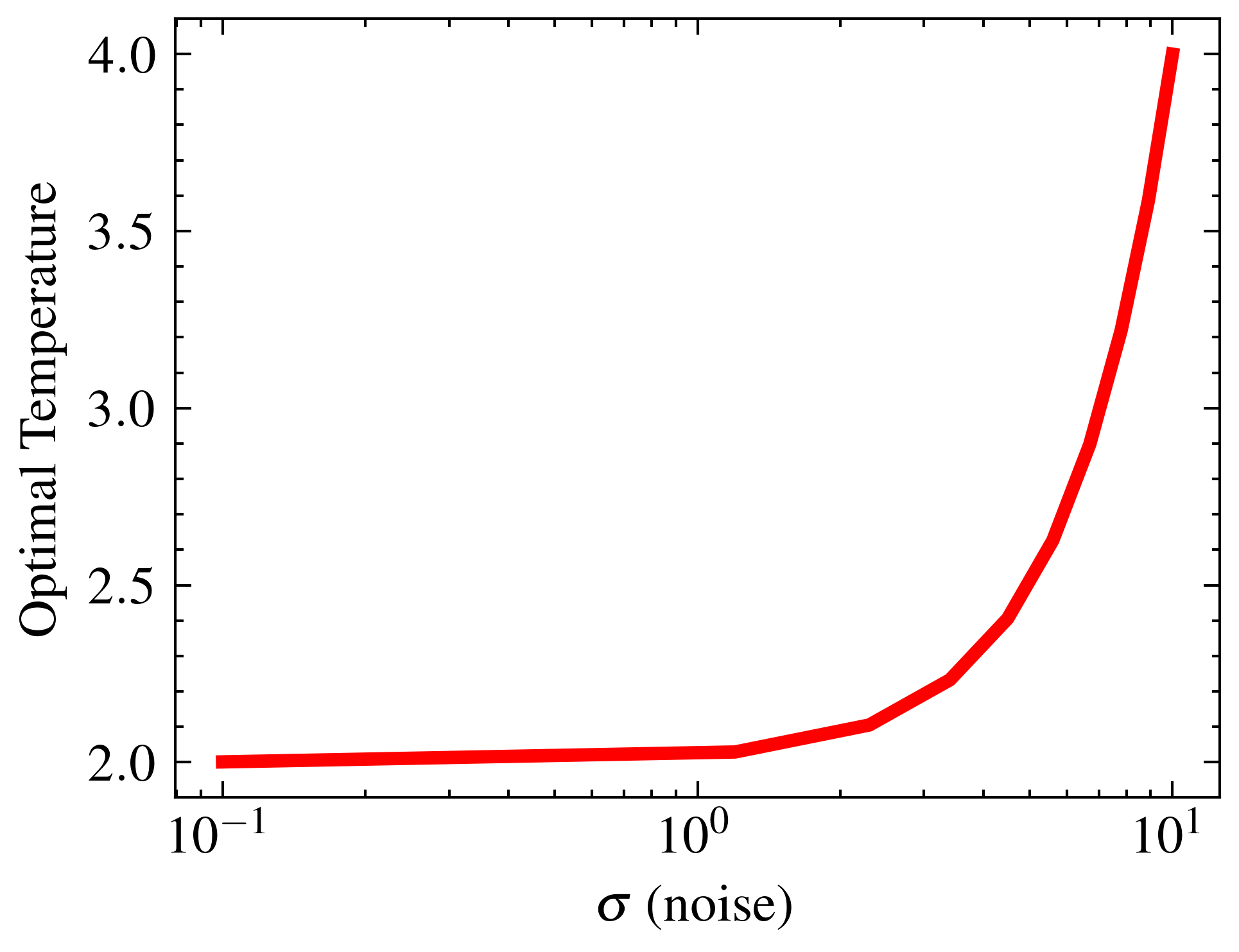}
         \captionsetup{justification=centering}
         \caption{Optimal temperature}
     \end{subfigure}
     \caption{Effect of noise shift on Transformer \eqref{eq:approximate_attention}. The pretraining noise is $\sigma_{train} = 0.1$, while $\sigma_{test}$ varies across plots. The optimal temperature is set by Theorem~\ref{theorem:optimal_temperature}. This setting matches Figure~\ref{fig:approximate_attention_experiments}a, except for changes in test-time noise $\sigma_{test}$.}
     \label{fig:noise_shift}
\end{figure*}

\subsection{Effect of distribution shift}

In this section, we explore scenarios where $\mathcal{D}^{test} \neq \mathcal{D}^{train}$, indicating a shift in the input, task, or noise distribution after pretraining the model. We consider three cases: (1) a shift in the input distribution (altered mean or covariance), (2) a shift in the task distribution, and (3) a change in the noise levels.

\paragraph{Pretraining for explaining distribution shift effects ---} To streamline our explanations below, for pretraining, we optimize the parameters $\mV$ and $\mM$ using $m$ samples of $(\mZ, y_l)$ drawn from the distribution $\mathcal{D}^{train}$, where each $\mZ$ contains $l-1$ $(\vx, y)$ pairs intended for ICL. Building upon prior work that connects ICL in linear regression to the Bayes-optimal ridge estimator \citep{zhang2024trained, raventos2024pretraining}, we configure $\mM$ and $\mV$ to emulate Bayes-optimal ridge regression. Specifically, we aim for $\hat{\vw}_{Att}(\mC_{xx}, \mC_{xy}; \mM, \mV) \approx \hat{\vw}_{Bayes}$ and $b_{Att}(\vs_x, s_y; \mV) \approx 0$.

\begin{proposition}[Pretrained parameters]
When the temperature parameter is set to $\tau = 1$ during pretraining, the following parameter configuration approximates the Bayes-optimal estimator in \eqref{eq:bayes_optimal_ridge_estimator}:
\begin{align} \label{eq:pretraining}
    \mM_{11} &= d\left(\frac{\hat{\mX}^T \hat{\mX}}{ml} + \frac{\sigma^2}{l} \mSigma_w^{-1}\right)^{-1}, \quad \;\; \vm_{21} = \mathbf{0}, \\
    \vv_{21} &= \frac{\sigma^2}{dl} \left( \frac{\hat{\mX}^T \hat{\mX}}{ml}  \right)^{-1} \mSigma_w^{-1} \vmu_w, \quad \quad v_{22} = \frac{1}{d},\nonumber
\end{align}
where $\hat{\mX} \in \mathbb{R}^{ml \times d}$ is the centered input matrix formed from $ml$ samples of $\vx$. This configuration aligns the our model with Bayes-optimal ridge regression. The quantities $\vmu_w$ and $\mSigma_w$ can be estimated from the pretraining data. A detailed derivation is provided in Appendix \ref{appendix:pretraining}.
\label{lemma:pretraining}
\end{proposition}

\begin{remark}
    While the pretrained parameters specified in Proposition~\ref{lemma:pretraining} are not guaranteed to be optimal in all settings, they are analytically useful for examining the effects of distribution shifts.
\end{remark}

Based on Proposition \ref{lemma:pretraining}, we arrive at the following:
\begin{corollary}
    Suppose there is no distribution shift between training and inference. Then, under the parameter configuration of Proposition \ref{lemma:pretraining}, the Transformer model \eqref{eq:approximate_attention} emulates the Bayes-optimal linear model, implying that it is capable of in-context learning according to Definition \ref{definition:icl}.
\end{corollary}

Since the pretrained model succeeds in ICL for $\mathcal{D}^{test} = \mathcal{D}^{train}$, we next investigate how distribution shifts affect its ICL performance.

\paragraph{ICL under distribution shift ---} To evaluate the impact of these distribution shifts on ICL performance, we assess whether adjustments to $\mM$ and/or $\mV$ are necessary to match the Bayes-optimal linear model under the new distribution. If so, the model is considered sensitive to the shift. Otherwise, it is deemed robust.

\paragraph{Case I: Shift in input distribution ---}
Recall that inputs are drawn as $\vx_i \sim \mathcal{N}(\vmu_x, \mSigma_x)$, as defined in \eqref{eq:data_model}. Let $\vmu_x^{train}, \mSigma_x^{train}$ and $\vmu_x^{test}, \mSigma_x^{test}$ denote the input means and covariances for pretraining and testing, respectively. We consider two subcases:

\begin{enumerate}[itemsep=0cm,parsep=0cm,leftmargin=0.5cm,label=(\roman*)]
    \item {Mean shift ($\vmu_x^{train}\neq\vmu_x^{test}$):} Centering renders the approximate model invariant to mean shifts, but the uncentered linear attention model remains sensitive, as noted in Remark~\ref{remark:linear_vs_approximate}.
    \item {Covariance shift ($\mSigma_x^{train}\neq\mSigma_x^{test}$):} 
Since $\mM_{11}$ is fitted to the pretraining covariance, a mismatch drives the estimator away from Bayes-optimality, echoing prior results on linear attention~\citep{zhang2024trained}.
\end{enumerate}

\paragraph{Case II: Shift in task distribution ---}
The task vectors follow $\vw \sim \mathcal{N}(\vmu_w, \mSigma_w)$. Let $\vmu_w^{train}, \mSigma_w^{train}$ and $\vmu_w^{test}, \mSigma_w^{test}$ be the mean and covariance of the task distribution during pretraining and testing, respectively. The Transformer model can incorporate $\vmu_w^{train}$ and $\mSigma_w^{train}$ via the pretrained parameters $\mM_{11}$ and $\vv_{21}$ (see Proposition~\ref{lemma:pretraining}). However, as the context length $l$ increases, the model’s dependence on the task distribution diminishes. Thus, shifts in the task distribution primarily affect ICL performance for small $l$.

\paragraph{Case III: Shift in noise distribution ---}

Finally, consider a change in the noise distribution: $\epsilon_i \sim \mathcal{N}(0, \sigma^2)$, with $\sigma^2_{train}$ and $\sigma^2_{test}$ denoting pretraining and testing noise variances. If $\sigma^2_{train} \neq \sigma^2_{test}$, the parameters $\mM_{11}$ and $\vv_{21}$ become suboptimal relative to the Bayes-optimal linear model. However, as with the task distribution, the impact of noise shift diminishes as $l \to \infty$.

\noindent\textbf{Summary ---} The Transformer model is robust to shifts in input mean but sensitive to input covariance changes. Shifts in task or noise distribution reduce ICL performance at small $l$, though increasing $l$ mitigates these effects. 

\subsection{Interpretation of the optimal attention temperature (Theorem \ref{theorem:optimal_temperature})}
\label{section:interpretation_optimal_temperature}
To obtain a simplified expression, we consider a simple but representative family of shifts as follows. The training distributions are
\[
\vx_i \sim \mathcal{N}(\mathbf{0}, \mI), \qquad \vw \sim \mathcal{N}(\mathbf{0}, \mI), \qquad \epsilon_i \sim \mathcal{N}(0, \hat{\sigma}^2).
\]
We then introduce three independent shift parameters for the test distribution: 
\[
\vx_i \sim \mathcal{N}(\mathbf{0}, a \mI), \qquad \vw \sim \mathcal{N}(\mathbf{0}, b \mI), \qquad \epsilon_i \sim \mathcal{N}(0, \sigma^2),
\]
where $a>0$ controls the input variance shift, $b>0$ controls the task-parameter variance shift, and $\sigma>0$ controls the noise-variance shift. This setting preserves isotropy, which makes it possible to derive a clean closed-form expression while still connecting directly to realistic distribution shifts.

Substituting these shifted distributions into the optimal-temperature expression in \eqref{eq:optimal_temperature} yields
\begin{align}
    \tau_{\text{opt}} &= \frac{2\text{Tr}\left(a \mI \mM_{11}^T \left(ab \mI + \frac{1}{l} (\sigma^2 + ab d) \mI \right) a \mI \mM_{11} \right)}{\text{Tr}\left(a \mI \left(ab \mI \left( \mM_{11} + \mM_{11}^T \right)\right)\right)},\\
    &= \left(a + \frac{1}{l}\left( \frac{\sigma^2}{b} + ad \right) \right) \frac{\text{Tr}\left( \mM_{11}^T \mM_{11} \right)}{ \text{Tr}\left(\mM_{11}\right)}. \label{eq:derived_example_optimal_temperature}
\end{align}

This concrete formula makes several effects fully explicit:

\begin{itemize}[leftmargin=*]
    \item \textit{Input shift ---} Increasing input variance $a$ directly scales $\tau_{\text{opt}}$ upward. This aligns with our earlier results (Figure \ref{fig:approximate_attention_experiments}) and the heuristic derived in Appendix J, which suggests that a greater variance of pre-softmax scores requires a higher temperature to maintain robustness to input shifts.
    \item \textit{Noise shift ---} Increasing noise variance $\sigma^2$ also increases $\tau_{\text{opt}}$, but only through the $\tfrac{1}{l}$ term, reflecting the diminishing effect of noise when more in-context examples are available.
    \item \textit{Task shift ---} Increasing task variance $b$ reduces the effect of noise (via $\sigma^2/b$), slightly lowering the optimal temperature.
    \item \textit{Context length ---} As $l\to\infty$, the $\tfrac{1}{l}$ term vanishes, giving a simplified asymptotic rule: $\tau_{\text{opt}} \to a \cdot 
        \frac{\text{Tr}(\mM_{11}^\top \mM_{11})}{\text{Tr}(\mM_{11})}.$
\end{itemize}
Note that under the considered training distribution, we have $\mathrm{Tr}(\mM_{11}^\top \mM_{11}) / \mathrm{Tr}(\mM_{11}) \approx 1$, which implies that $\tau_{\mathrm{optimal}} \to a$ as $l \to \infty$.

\subsection{Possible extensions to the general softmax setting: the moment-ratio heuristic}

We propose a practical heuristic (derived in Appendix~\ref{appendix:insights_for_other_settings}): the optimal attention temperature is roughly proportional to the ratio of the second and first moments of pre-softmax attention scores. The expression in \eqref{eq:derived_example_optimal_temperature} provides further theoretical justification for that heuristic.

Indeed, for some $i\neq j$,
\begin{align}
    \tau_{\text{opt}} &= a \frac{\text{Tr}\left( \mM_{11} \mM_{11}^T \right)}{ \text{Tr}\left(\mM_{11}\right)} + \frac{1}{l}\left( \frac{\sigma^2}{b} + ad \right) \frac{\text{Tr}\left( \mM_{11}\mM_{11}^T \right)}{ \text{Tr}\left(\mM_{11}\right)},\\
    &= \frac{\E\!\left[(\vz_i^{\top}\mM\vz_j)^2\right]}{\E\!\left[\vz_i^{\top}\mM\vz_i\right]} + \frac{1}{l}\left( \frac{\sigma^2}{b} + ad \right) \frac{\text{Tr}\left( \mM_{11} \mM_{11}^T \right)}{ \text{Tr}\left(\mM_{11}\right)},
\end{align}
where we used the moments calculated in Appendix \ref{appendix:insights_for_other_settings} to reach the final line.
Here, since $\mathrm{Tr}(\mM_{11}^\top \mM_{11}) / \mathrm{Tr}(\mM_{11}) \approx 1$ for the considered training distribution, this gives the approximation:
\begin{align}
    \tau_{\text{opt}} \approx \underbrace{\frac{\E\!\left[(\vz_i^{\top}\mM\vz_j)^2\right]}{\E\!\left[\vz_i^{\top}\mM\vz_i\right]}}_{\text{moment-ratio}} + \underbrace{\frac{1}{l}\left( \frac{\sigma^2}{b} + ad \right)}_{\text{correction for small $l$}}.\label{eq:moment_ratio_with_correction}
\end{align}

This demonstrates that the moment-ratio heuristic is not an ad-hoc rule, but a theoretically grounded approximation of the exact closed-form optimal temperature.

\begin{figure*}[t]
    \centering
    \begin{subfigure}[b]{0.49\textwidth}
         \centering
         \includegraphics[width=0.99\linewidth]{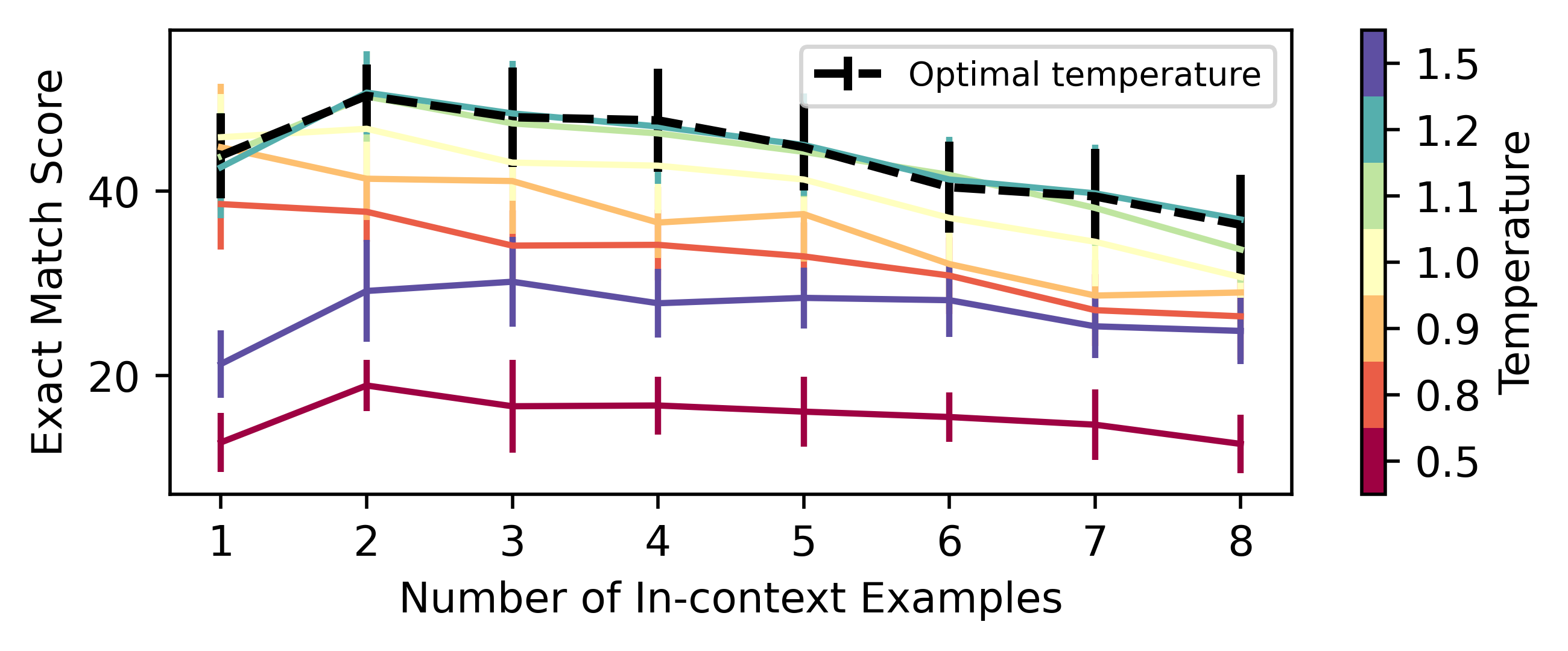}
         \caption{Effect of context length}
     \end{subfigure}
      \begin{subfigure}[b]{0.49\textwidth}
         \centering
         \includegraphics[width=0.99\linewidth]{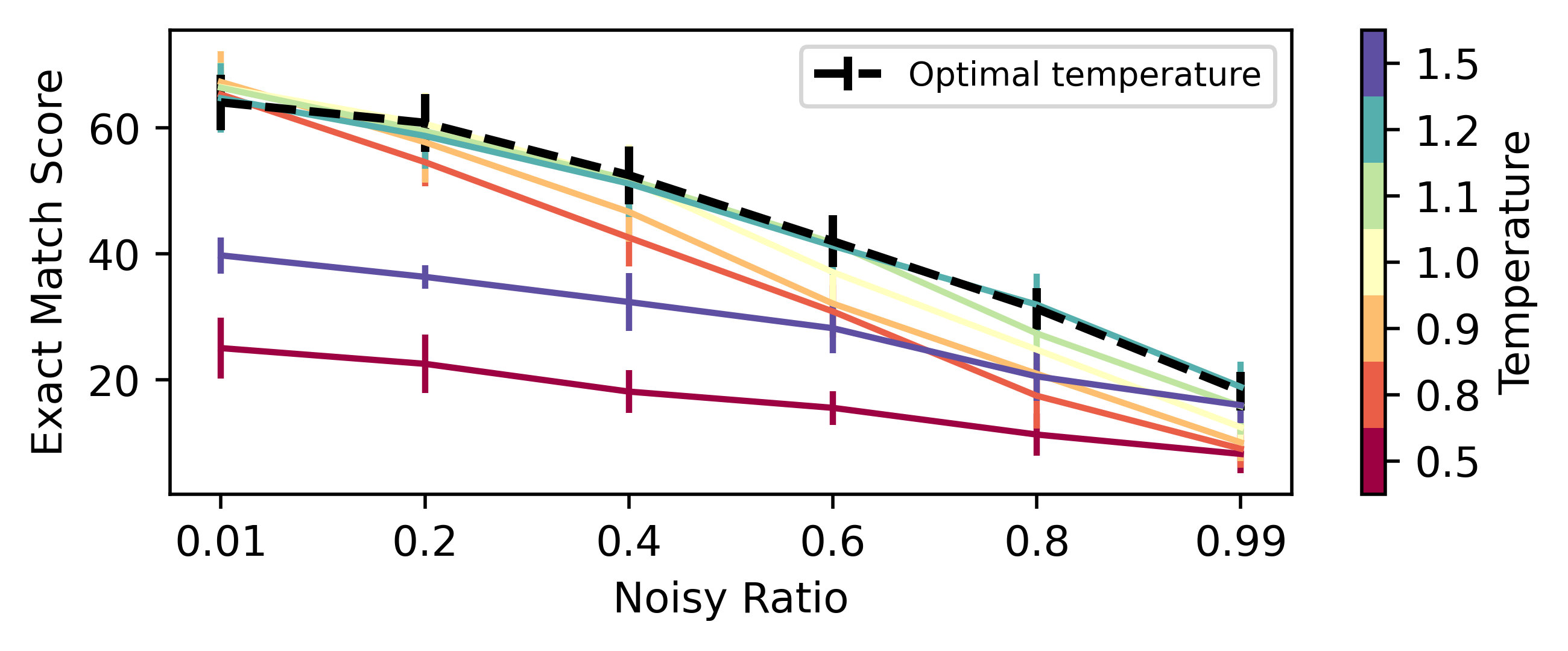}
         \caption{Effect of noisy ratio}
     \end{subfigure}
     \caption{Effect of attention temperature on the ICL performance of Llama2-7B \citep{touvron2023llama} on the SCIQ dataset \citep{welbl2017crowdsourcing}. 
Distribution shift is induced by injecting noisy yet “relevant” labels into in-context demonstrations following \citet{gao2024on}. Panel (a) fixes the noisy ratio at 0.6; panel (b) fixes the number of in-context examples at 6. Results (averaged over 12 Monte Carlo runs) include error bars showing one standard deviation. Attention temperature of all the layers is set to $\tau\sqrt{d_k}$ for dimension independence, where $d_k$ denotes the key dimension of the corresponding layer. Furthermore, the dashed black line marks the “optimal temperature” computed from the variance-to-mean ratio of pre-softmax scores, which is an insight derived from Theorem~\ref{theorem:optimal_temperature}, as explained in Appendix~\ref{appendix:insights_for_other_settings}. Full experimental details appear in Appendix~\ref{appendix:experimental_details}.
}

     \label{fig:LLM}
\end{figure*}

\vspace{-1em}
\section{Experimental results}

In this section, we empirically validate our theory and show that optimal attention temperature consistently enhances generalization. We begin with controlled linear regression experiments using (i) the simplified Transformer model with approximate softmax attention \eqref{eq:approximate_attention} and (ii) GPT-2 \citep{radford2019language}, which combines multi-head softmax attention with MLP layers\footnote{GPT-2 results are in Appendix~\ref{appendix:experimental_details}.}. These experiments confirm that our theoretical insights transfer from simplified to expressive architectures. Finally, we evaluate Llama2-7B \citep{touvron2023llama} on SCIQ in-context learning tasks \citep{welbl2017crowdsourcing}, demonstrating that temperature selection is a principled and effective lever for improving robustness in large language models.
\vspace{-1em}
\subsection{Experiments on linear regression tasks}
\vspace{-0.5em}
We consider a Transformer architecture with approximate softmax attention and no MLP layers, as analyzed in our theoretical development. Figures~\ref{fig:approximate_attention_experiments} and~\ref{fig:noise_shift} illustrate its behavior on linear regression tasks \eqref{eq:data_model}. Theoretical predictions closely match empirical performance across a range of conditions, confirming the robustness of our analysis. In Figure~\ref{fig:approximate_attention_experiments}, we compare the ICL performance of the model with and without applying the optimal temperature. As context length $l$ increases (Figure~\ref{fig:approximate_attention_experiments}a), the model's predictions converge to those of the Bayes-optimal linear model, validating its ICL capability. Figure~\ref{fig:approximate_attention_experiments}b shows that under an input covariance shift, model performance degrades—but applying the optimal temperature restores alignment with the Bayes-optimal solution. Additionally, Figure~\ref{fig:approximate_attention_experiments}c shows that the influence of task distribution shift decreases as $l$ increases.

We further evaluate robustness to label noise in Figure~\ref{fig:noise_shift}. In Figure~\ref{fig:noise_shift}a, we observe that noise effects diminish as the context length increases, consistent with our theoretical predictions. However, at small $l$, temperature adjustment becomes critical. In Figure~\ref{fig:noise_shift}b (for $l = d$), the Transformer increasingly diverges from the Bayes-optimal model as noise grows, yet optimal temperature correction closes this gap. Figure~\ref{fig:noise_shift}c shows that the optimal temperature increases with noise level, indicating a principled relationship between noise and temperature under limited context.

\vspace{-1em}
\subsection{Experiments with LLMs for in-context Q\&A}

To assess the practical relevance of our theoretical framework, we investigate how attention temperature impacts the ICL behavior of LLMs. Since the optimal temperature in Theorem \ref{theorem:optimal_temperature} is not directly applicable here due to setting differences, we derive insights regarding temperature choice in other settings based on the optimal temperature in Theorem \ref{theorem:optimal_temperature}. Specifically, the insight is that the temperature choice should be proportional to the ratio of the variance of pre-softmax scores to the mean of those, which is described in Appendix \ref{appendix:insights_for_other_settings} in detail. 

Following \citet{gao2024on}, we generate SCIQ-based \citep{welbl2017crowdsourcing} ICL tasks that introduce distribution shift via noisy labels, with prompt examples and label construction detailed in Appendix~\ref{appendix:experimental_details}. We evaluate Llama2-7B \citep{touvron2023llama} using exact-match score. 

Figure~\ref{fig:LLM} shows the results. In (a) and (b), optimal attention temperature enhances the ICL performance for various context lengths and noisy ratios (analogous to Figures \ref{fig:noise_shift}a and \ref{fig:noise_shift}b). Furthermore, in (b), higher noise ratios push the optimal temperature upward, matching our theoretical prediction (cf. Figure~\ref{fig:noise_shift}c). Together, these experiments demonstrate that the optimal temperature is not only theoretically motivated but also an effective tool for improving ICL robustness in real-world LLMs.

\vspace{-1em}
\section{Conclusion}

This work provides a unified theoretical and empirical account of how attention temperature governs the in-context learning (ICL) performance of pretrained Transformers under distribution shift.  
Using a simplified yet expressive framework based on \emph{approximate softmax attention}, we analytically show how shifts in input covariance and label noise degrade ICL and derive an \emph{optimal temperature} that provably minimizes generalization error.  
Extensive experiments on synthetic regression tasks, GPT-2, and Llama-2 validate our predictions, demonstrating that temperature selection is not a mere heuristic but a principled mechanism for improving robustness. Taken together, our results advance the theoretical understanding of Transformer behavior under distribution shift and establish attention temperature as a powerful, practical lever for building more adaptive and generalizable foundation models.

\vspace{-1em}
\section*{Impact Statement}
This paper presents work whose goal is to advance the field of Machine
Learning. There are many potential societal consequences of our work, none
which we feel must be specifically highlighted here.

\section*{Acknowledgements}
This work is supported in part by TÜBİTAK under project 124E063 within the ARDEB 1001 program, and we gratefully acknowledge their support. S.D. is further supported by an AI Fellowship from the KUIS AI Center and a PhD Scholarship (BİDEB 2211) from TÜBİTAK.

\bibliography{ref}
\bibliographystyle{icml2026}

\newpage
\appendix
\onecolumn

\section{Extended discussion of the related work}
\label{appendix:related_work}

\paragraph{ICL by Transformers ---} The ICL capability of Transformers was first brought to prominence by \cite{brown2020language}, leading to a surge of empirical and theoretical investigations. Several works have demonstrated that ICL performance improves with model scale \cite{wei2022emergent, olsson2022context, schaeffer2023are}, underscoring its importance in modern AI systems. To better understand this phenomenon, synthetic tasks such as linear regression have served as controlled testbeds for analyzing ICL in Transformers \citep{garg2022can, zhang2024trained, raventos2024pretraining}. A prevailing hypothesis in recent theoretical work is that Transformers implicitly learn algorithms during pretraining, which they subsequently execute during inference \citep{bai2023transformers, li2023transformers, akyurek2023what, ahn2023transformers, pmlr-v202-von-oswald23a, mahankali2024one, fu2024transformers, zhang2024trained, li2024finegrained, park2024competition}. There remains ongoing debate over the precise nature of these learned procedures. However, our work focuses on a fundamentally different question, which is how attention temperature affects the ICL performance of pretrained Transformers under distribution shifts.

\section{Derivation of Bayes-optimal ridge estimator for \texorpdfstring{$\vw$}{w}}
\label{appendix:bayes_optimal}

We derive the Bayes-optimal ridge estimator for $\vw$ given a set of context samples.  
We place a Gaussian prior on $\vw$, assumed to be a random vector $\vw \sim \mathcal{N}(\vmu_0, \mSigma_0)$ with prior mean $\vmu_0$ and covariance $\mSigma_0$.  
Let the observed (centered) inputs and labels be
\[
\bar{\mX} = [\bar{\vx}_1,\dots,\bar{\vx}_{l-1}]^T, \quad
\bar{\vy} = [\bar{y}_1,\dots,\bar{y}_{l-1}]^T,
\]
and assume i.i.d. Gaussian noise $\epsilon_i\sim\mathcal{N}(0,\sigma^2)$.  
The likelihood of $\bar{\vy}$ given $\vw$ is
\begin{align}
    p(\bar{\vy}\mid\bar{\mX},\vw)
    &= \prod_{i=1}^{l-1}\frac{1}{\sqrt{2\pi\sigma^2}}\exp\!\left[-\frac{(\bar{y}_i-\vw^T\bar{\vx}_i)^2}{2\sigma^2}\right] \\
    &\propto \exp\!\left[-\frac{1}{2\sigma^2}(\bar{\vy}-\bar{\mX}\vw)^T(\bar{\vy}-\bar{\mX}\vw)\right],
\end{align}
where $\propto$ denotes proportionality.

By Bayes’ rule, the posterior of $\vw$ is proportional to the product of likelihood and prior:
\begin{align}
    p(\vw\mid\bar{\vy},\bar{\mX})
    \propto
    p(\bar{\vy}\mid\bar{\mX},\vw)\,p(\vw).
\end{align}
Substituting the Gaussian prior yields
\begin{align}
    p(\vw\mid\bar{\vy},\bar{\mX}) \propto
    \exp\!\Big[-\frac{1}{2\sigma^2}(\bar{\vy}-\bar{\mX}\vw)^T(\bar{\vy}-\bar{\mX}\vw)\Big]
    \exp\!\Big[-\frac{1}{2}(\vw-\vmu_0)^T\mSigma_0^{-1}(\vw-\vmu_0)\Big].
\end{align}

To determine the form of the posterior distribution, we complete the square in the exponent by collecting all terms involving $\vw$. Expanding the exponent in the joint expression from above, we obtain:
\begin{align}
    -\frac{1}{2\sigma^2} \left(\bar{\vy}^T \bar{\vy} - 2\bar{\vy}^T \bar{\mX} \vw + \vw^T \bar{\mX}^T \bar{\mX} \vw\right)
    - \frac{1}{2} \left(\vw^T \mSigma_0^{-1} \vw - 2\vmu_0^T \mSigma_0^{-1} \vw + \vmu_0^T \mSigma_0^{-1} \vmu_0\right).
\end{align}

Grouping the quadratic and linear terms in $\vw$, we arrive at:
\begin{align}
    -\frac{1}{2} \vw^T \left( \frac{\bar{\mX}^T \bar{\mX}}{\sigma^2} + \mSigma_0^{-1} \right) \vw
    + \vw^T \left( \frac{\bar{\mX}^T \bar{\vy}}{\sigma^2} + \mSigma_0^{-1} \vmu_0 \right)
    + \text{terms independent of } \vw.
\end{align}

Defining the posterior precision and linear coefficient terms as $\mSigma_l^{-1} = \frac{\bar{\mX}^T \bar{\mX}}{\sigma^2} + \mSigma_0^{-1}$ and $b_l = \frac{\bar{\mX}^T \bar{\vy}}{\sigma^2} + \mSigma_0^{-1} \vmu_0$, the exponent can be rewritten as
\begin{align}
    -\frac{1}{2} \vw^T \mSigma_l^{-1} \vw + \vw^T b_l 
    = -\frac{1}{2} (\vw - \vmu_l)^T \mSigma_l^{-1} (\vw - \vmu_l) + \text{const},
\end{align}
where $\vmu_l = \mSigma_l b_l$ denotes the posterior mean. Expanding this expression gives:
\begin{align}
    \vmu_l = \left(\frac{\bar{\mX}^T \bar{\mX}}{\sigma^2} + \mSigma_0^{-1} \right)^{-1}
             \left( \frac{\bar{\mX}^T \bar{\vy}}{\sigma^2} + \mSigma_0^{-1} \vmu_0 \right).
\end{align}

Hence, the posterior distribution of $\vw$ given the observed data is Gaussian:
\begin{align}
    \vw \mid \bar{\vy}, \bar{\mX} \sim \mathcal{N}(\vmu_l, \mSigma_l),
\end{align}
where $\vmu_l$ is the posterior mean and $\mSigma_l$ is the posterior covariance matrix.

Under squared-error loss, the Bayes-optimal estimator coincides with the posterior mean, yielding the Bayes-optimal ridge estimator:
\begin{align}
    \hat{\vw}_{\text{Ridge}}=\mathbb{E}[\vw\mid\bar{\vy},\bar{\mX}]=\vmu_l
    =\Big(\frac{\bar{\mX}^T\bar{\mX}}{\sigma^2}+\mSigma_0^{-1}\Big)^{-1}
      \Big(\frac{\bar{\mX}^T\bar{\vy}}{\sigma^2}+\mSigma_0^{-1}\vmu_0\Big).
\end{align}

This expression provides the Bayes-optimal ridge estimate of $\vw$ under a Gaussian prior and additive Gaussian noise—minimizing expected squared error with respect to the posterior.

\section{Derivation of approximate softmax}
\label{appendix:approximate_softmax}

The function $\text{softmax} : \R^l \to \R^l$ is defined component-wise as
\begin{align}
    \text{softmax}(\vz)_i := \frac{e^{z_i}}{\sum_{j=1}^{l} e^{z_j}} \quad \forall i \in \{1, \dots, l\}.
\end{align}
To obtain an approximation, we expand around the origin $\vz = \mathbf{0}$ using a first-order Taylor series:
\begin{align}
    \text{softmax}(\vz) \approx \text{softmax}(\mathbf{0}) + J_{\text{softmax}}(\mathbf{0}) \vz,
\end{align}
where $J_{\text{softmax}}(\mathbf{0})$ is the Jacobian matrix of the softmax function evaluated at $\vz = \mathbf{0}$.  

We first compute the zeroth-order term:
\begin{align}
    \text{softmax}(\mathbf{0}) = \frac{e^{0}}{\sum_{j=1}^{l} e^{0}} \mathbf{1} = \frac{1}{l} \mathbf{1}.
\end{align}

Next, we evaluate the Jacobian entries at $\vz = \mathbf{0}$:
\begin{align}
    J_{\text{softmax}}(\mathbf{0})_{ii} &= \text{softmax}(\mathbf{0})_i \left(1 - \text{softmax}(\mathbf{0})_i\right) = \frac{l - 1}{l^2}, \quad \forall i, \\
    J_{\text{softmax}}(\mathbf{0})_{ij} &= -\text{softmax}(\mathbf{0})_i \cdot \text{softmax}(\mathbf{0})_j = -\frac{1}{l^2}, \quad \forall i \neq j.
\end{align}
This yields the compact matrix form:
\begin{align}
    J_{\text{softmax}}(\mathbf{0}) = \frac{1}{l} \mI - \frac{1}{l^2} \mathbf{1}\mathbf{1}^T.
\end{align}

Substituting back, we obtain the approximate softmax:
\begin{align}
    \text{softmax}(\vz) &\approx \frac{1}{l} \mathbf{1} + \left(\frac{1}{l} \mI - \frac{1}{l^2} \mathbf{1}\mathbf{1}^T\right)\vz, \\
                        &= \left(\frac{1}{l} - \frac{1}{l^2} \sum_{j=1}^l z_j \right)\mathbf{1} + \frac{1}{l}\vz, \\
                        &=: \widehat{\text{softmax}}(\vz).
    \label{eq:approximate_softmax_definition}
\end{align}

This derivation yields the approximate softmax attention formulation in \eqref{eq:approximate_attention}. From a practical standpoint, approximate softmax attention mechanisms have been empirically evaluated and shown to achieve performance comparable to standard softmax attention \citep{han2024bridging}.

\section{Temperature effects for softmax and approximate softmax}
\label{appendix:temperature_in_approximate_softmax}
The temperature parameter in softmax directly controls the variance of the output distribution. At higher temperatures, the variance across components decreases, and in the limit $\tau \to \infty$, all elements converge to $1/l$ with zero variance. Conversely, lower temperatures increase variance, and as $\tau \to 0^+$, the output approaches a one-hot vector, achieving maximal variance. 

In the approximate case, temperature similarly acts as an inverse scaling of the variance of the output components, capturing the limit $\tau \to \infty$ (all elements equal to $1/l$). For $\tau \to 0^+$, approximate softmax also reflects the maximal variance, but it does not produce a true one-hot distribution. Thus, approximate softmax closely mirrors the temperature behavior of softmax, except in the degenerate limit $\tau \to 0^+$, which is not of practical relevance in this work. 

To further illustrate these effects, Figure~\ref{fig:temperature_effects_approximate_vs_softmax} compares softmax and approximate softmax across different temperatures. The figure demonstrates that approximate softmax faithfully captures the variance effect of temperature: the variance of the output components is inversely proportional to $\tau$. Moreover, as $\tau \to \infty$, both softmax and approximate softmax concentrate around $1/l$, whereas linear attention with temperature scaling does not. Overall, the output distributions of softmax and approximate softmax are highly similar, except at very small values of $\tau$, where approximate softmax may yield negative components while softmax tends toward sparsity with many zeros. By contrast, linear attention with temperature scaling produces qualitatively different distributions. This comparison highlights the advantage of approximate softmax as a faithful surrogate for analyzing temperature effects relevant to softmax.

\section{Expanded form of approximate softmax attention}
\label{appendix:expanded_approximate_attention}

Using block matrix notation, the prediction from the approximate softmax attention model can be expanded as:
\begin{align}
    &\hat{y}(\mZ;\mV, \mM) = A_{d+1, l},\\
    &=\frac{1}{l} \mV_{d+1,:} \mZ \left(\frac{(\mK \mZ)^T (\mQ \mZ_{:,l})}{\tau}  - \frac{\mathbf{1}}{l} \sum_{j=1}^{l} \frac{(\mK \mZ_{:,j})^T (\mQ \mZ_{:,l})}{\tau} + \mathbf{1} \right),\\
    &= \frac{1}{l}  [ \vv_{21}^T \;\; v_{22}] \mZ \left(\frac{\mZ^T \mM \mZ_{:,l}}{\tau}  - \frac{\mathbf{1}}{l} \sum_{j=1}^{l} \frac{(\mZ_{:,j})^T \mM \mZ_{:,l}}{\tau} + \mathbf{1} \right),\\
    &= \frac{1}{l}  [ \vv_{21}^T \;\; v_{22}] [\mX \;\; \vy]^T \left(\frac{[\mX \;\; \vy] \mM [\vx_l^T \;\; 0]^T}{\tau}  - \frac{\mathbf{1}}{l} \sum_{j=1}^{l} \frac{[\vx_i^T \;\; y_i] \mM [\vx_l^T \;\; 0]^T}{\tau} + \mathbf{1} \right),\\
    &= \frac{1}{l}  [ \vv_{21}^T \;\; v_{22}] [\mX \;\; \vy]^T \left(\frac{1}{\tau} \left[\mX - \mathbf{1}\vs_x^T  \;\; \vy - s_y \mathbf{1} \right] \begin{bmatrix} \mM_{11} & *\\ \vm_{21}^T & * \end{bmatrix} [\vx_l^T \;\; 0]^T  + \mathbf{1} \right),\\
    &= \frac{1}{l}  [ \vv_{21}^T \;\; v_{22}]  [\mX \;\; \vy]^T \left(\frac{1}{\tau} \left(\mX - \mathbf{1}\vs_x^T \right) \mM_{11} \vx_l  + \frac{1}{\tau} \left(\vy - s_y \mathbf{1} \right) \vm_{21}^T \vx_l  + \mathbf{1} \right),\\
    &= \frac{1}{l}  \left( \vv_{21}^T \mX^T +  v_{22} \vy^T \right) \left(\frac{1}{\tau} \left(\mX - \mathbf{1}\vs_x^T \right) \mM_{11} \vx_l  + \frac{1}{\tau} \left(\vy - s_y \mathbf{1} \right) \vm_{21}^T \vx_l  + \mathbf{1} \right),\\
    &= \frac{1}{\tau} \left( \vv_{21}^T \left( \frac{\mX^T \mX}{l} - \vs_x \vs_x^T \right) + v_{22} \left( \frac{\vy^T\mX}{l} - s_y \vs_x^T \right) \right) \mM_{11} \vx_l, \nonumber\\
    & \qquad \qquad + \frac{1}{\tau} \left( \vv_{21}^T \left( \frac{\mX^T \vy}{l} - s_y \vs_x \right) + v_{22} \left( \frac{\vy^T \vy}{l} - s_y^2 \right) \right) \vm_{21}^T \vx_l + \vv_{21}^T \vs_x + v_{22} s_y,\\
    &= \frac{1}{\tau} \left( \vv_{21}^T \mC_{xx} + v_{22} \mC_{xy}^T \right) \mM_{11} \vx_l + \frac{1}{\tau} \left( \vv_{21}^T \mC_{xy} + v_{22} C_{yy} \right) \vm_{21}^T \vx_l + \vv_{21}^T \vs_x + v_{22} s_y,\\
    &= \frac{1}{\tau} \left( \left( \vv_{21}^T \mC_{xx} + v_{22} \mC_{xy}^T \right) \mM_{11} + \left( \vv_{21}^T \mC_{xy} + v_{22} C_{yy} \right) \vm_{21}^T \right) \vx_l + \vv_{21}^T \vs_x + v_{22} s_y,
\end{align}
where the summary statistics are defined as:
\begin{align*}
    \vs_x &:= \frac{1}{l} \sum_{i = 1}^{l} \vx_i, \qquad \qquad \qquad \;\;\quad
    s_y := \frac{1}{l} \sum_{i = 1}^{l-1} y_i, \\
    \mC_{xx} &:= \frac{1}{l} \sum_{i = 1}^{l} \vx_i \vx_i^T - \vs_x \vs_x^T,\qquad
    \mC_{xy} := \frac{1}{l} \sum_{i = 1}^{l-1} y_i \vx_i - s_y \vs_x, \qquad
    C_{yy} := \frac{1}{l} \sum_{i = 1}^{l-1} y_i^2 - s_y^2.
\end{align*}

Then, we define 
\begin{align}
    \hat{\vw}_{Att}(\mC_{xx}, \mC_{xy}, C_{yy}; \mM, \mV) &= \mM_{11}^T \left( \mC_{xx} \vv_{21} + v_{22} \mC_{xy} \right)  + \left( \vv_{21}^T \mC_{xy} + v_{22} C_{yy} \right) \vm_{21},\\
    b_{Att}(\vs_x, s_y; \mV) &= \vv_{21}^T \vs_x + v_{22} s_y,
\end{align}
which allows us to write
\begin{align}
    \hat{y}(\mZ;\mV, \mM) = \frac{1}{\tau} \hat{\vw}_{Att}(\mC_{xx}, \mC_{xy}, C_{yy}; \mM, \mV)^T \vx_l + b_{Att}(\vs_x, s_y; \mV).
\end{align}

\section{Derivation of the pretraining for ICL by mimicking the Bayes-optimal estimator}
\label{appendix:pretraining}
Here, we derive the pretraining of the approximate softmax attention model by mimicking the Bayes-optimal ridge estimator \eqref{eq:bayes_optimal_ridge_estimator}. The prediction of the approximate softmax attention model can be written as 
\begin{align}
    \hat{y}(\mZ;\mV, \mM) = \frac{1}{\tau} \hat{\vw}_{Att}(\mC_{xx}, \mC_{xy}, C_{yy}; \mM, \mV)^T \vx_l + b_{Att}(\vs_x, s_y; \mV),
\end{align}
which is derived in Appendix \ref{appendix:expanded_approximate_attention}. Furthermore, the Bayes-optimal ridge regression model's prediction is 
\begin{align}
    \hat{y}_{Bayes} = \hat{\vw}_{Bayes}^T \vx_l.
\end{align}
Therefore, we select the parameters $\mM$ and $\mV$ such that
\begin{align}
    \hat{\vw}_{Att}(\mC_{xx}, \mC_{xy}, C_{yy}; \mM, \mV) \approx \hat{\vw}_{Bayes}, \quad b_{Att}(\vs_x, s_y; \mV) \approx 0,  
\end{align}
which makes the prediction of the approximate softmax attention model approximately equal to that of the Bayes-optimal regression. Furthermore, we consider $\tau = 1$ for the pretraining. Let's first focus on $\hat{\vw}_{Att}(\mC_{xx}, \mC_{xy}, C_{yy}; \mM, \mV) $ as follows
\begin{align}
    &\hat{\vw}_{Att}(\mC_{xx}, \mC_{xy}, C_{yy}; \mM, \mV) \nonumber\\
    &\qquad \qquad = \left( \mM_{11}^T \left( \mC_{xx} \vv_{21} + v_{22} \mC_{xy} \right)  + \left( \vv_{21}^T \mC_{xy} + v_{22} C_{yy} \right) \vm_{21} \right),\\
    &\qquad \qquad = \left( \mM_{11}^T \left( \frac{\bar{\mX}^T \bar{\mX}}{l} \vv_{21} + v_{22} \frac{\bar{\mX}^T \bar{\vy}}{l} \right)  + \left( \vv_{21}^T \frac{\bar{\mX}^T \bar{\vy}}{l} + v_{22} \frac{\bar{\vy}^T \bar{\vy}}{l} \right) \vm_{21} \right).
    \label{eq:w_att_in_derivation}
\end{align}
To reach the last line, we use the fact that $\mC_{xx} := \mX^T \mX /l - \vs_x \vs_x^T = \bar{\mX}^T \bar{\mX}/l$, $\mC_{xy} := \mX^T \vy/l - s_y \vs_x = \bar{\mX}^T \bar{\vy}/l$ and $C_{yy} = \bar{\vy}^T \bar{\vy}/l$, where $\bar{\mX} := \mX - \vs_x^T$ and $\bar{\vy} := \vy - s_y$ denote centered input matrix and centered label vector. Now, recall that the Bayes-optimal ridge estimator is 
\begin{align}
    \hat{\vw}_{Bayes} = \left(\frac{\bar{\mX}^T \bar{\mX}}{\sigma^2} + \mSigma_w^{-1}\right)^{-1} \left(\frac{\bar{\mX}^T \bar{\vy}}{\sigma^2} + \mSigma_w^{-1} \vmu_w\right),
    \label{eq:w_bayes_in_derivation}
\end{align}
as derived in Appendix \ref{appendix:bayes_optimal}. 
Looking at equations \eqref{eq:w_bayes_in_derivation} and \eqref{eq:w_att_in_derivation} together, we can see that setting the parameters as follows would make $\hat{\vw}_{Att} = \hat{\vw}_{Bayes}$ hold
\begin{align}
    \mM_{11} = \frac{l}{\sigma^2}\left(\frac{\bar{\mX}^T \bar{\mX}}{\sigma^2} + \mSigma_w^{-1}\right)^{-1}, \quad \vv_{21} = \frac{\sigma^2}{l} \left( \frac{\bar{\mX}^T \bar{\mX}}{l} \right)^{-1} \mSigma_w^{-1} \vmu_w, \quad \vm_{21} = \mathbf{0}, \quad v_{22} = 1.
    \label{eq:ideal_dynamic_parameters}
\end{align}
However, while Bayes-optimal estimator $\hat{\vw}_{Bayes}$ is different for each sample, the attention model should be pretrained and fixed. Thus, we replace $\bar{\mX}^T \bar{\mX}$ in \eqref{eq:ideal_dynamic_parameters} with $\hat{\mX}^T \hat{\mX}/m$ as follows, where $\hat{\mX} \in \R^{ml \times d}$ is the centred input matrix including all the (pre)training data consisting of $ml$ samples. 
\begin{align}
    \mM_{11} = \frac{l}{\sigma^2}\left(\frac{\hat{\mX}^T \hat{\mX}}{m\sigma^2} + \mSigma_w^{-1}\right)^{-1}, \quad \vv_{21} = \frac{\sigma^2}{l} \left( \frac{\hat{\mX}^T \hat{\mX}}{ml}  \right)^{-1} \mSigma_w^{-1} \vmu_w, \quad \vm_{21} = \mathbf{0}, \quad v_{22} = 1.
    \label{eq:optimal_parameters_before_scaling}
\end{align}
In practice, the variance of noise $\sigma^2$, the mean $\vmu_w$, and covariance $\mSigma_w$ of the task vectors are unknown. Yet, we can use their estimates based on the (pre)training data. 

Now, we can focus on making $b_{Att}(\vs_x, s_y; \mV) \approx 0$ hold as follows
\begin{align}
    b_{Att}(\vs_x, s_y; \mV) &= \vv_{21}^T \vs_x + v_{22} s_y,
\end{align}
where $\vs_x$ and $s_y$ are based on data so we have no control over them. Instead, by using Assumptions \ref{assumption:mean_cov_bound} and \ref{assumption:jointly_diverge}, we can choose $\vv_{21}$ and $v_{22}$ such that $b_{Att} \to 0$ as $l,d \to \infty$. Note that Assumption \ref{assumption:mean_cov_bound} makes $\vv_{21}^T \vs_x + v_{22} s_y$ bounded with high probability for $\vv_{21}$ and $v_{22}$ given in \eqref{eq:optimal_parameters_before_scaling}. Therefore, multiplying $\vv_{21}, v_{22}$ given in \eqref{eq:optimal_parameters_before_scaling} with $1/d$ would make $b_{Att} \to 0$ as $d \to \infty$. To fix the impact of the multiplication for $\hat{\vw}_{Att}$, we can multiply $\mM_{11}$ with $d$ as well. So, by applying the mentioned multiplications, we reach the following pretrained parameters mimicking the Bayes-optimal regression model
\begin{align}
    \mM_{11} = \frac{dl}{\sigma^2}\left(\frac{\hat{\mX}^T \hat{\mX}}{m\sigma^2} + \mSigma_w^{-1}\right)^{-1}, \quad \vv_{21} = \frac{\sigma^2}{dl} \left( \frac{\hat{\mX}^T \hat{\mX}}{ml}  \right)^{-1} \mSigma_w^{-1} \vmu_w, \quad \vm_{21} = \mathbf{0}, \quad v_{22} = \frac{1}{d}.
    \label{eq:optimal_parameters}
\end{align}

\section{Characterization of generalization error for ICL under distribution shift}
\label{appendix:generalization_error}
Here, we characterize the generalization error for in-context learning under distribution shift, given that $\mM$ and $\mV$ are pretrained and fixed. So, the impact of pretraining distribution $\mathcal{D}^{train}$ is captured by $\mM$ and $\mV$. Suppose that $\mathcal{D}^{test}$ denotes the test distribution. To avoid additional notations, here, we again use $\vmu_x, \vmu_w, \mSigma_x, \mSigma_w, \sigma^2$ to denote means and covariances for input and task vectors and noise variance for the inference (test). However, note that these can be different from those used for pretraining. We begin studying the generalization error defined in \eqref{eq:generalization_definition} as follows
\begin{align}
    &\mathcal{G}(\mV, \mM) := \E_{(\mZ, y_{l}) \sim \mathcal{D}^{test}} \left[ \left( y_{l} - \hat{y}(\mZ;\mV, \mM) \right)^2  \right],\\
    &\quad = \E_{(\mZ, y_{l}) \sim \mathcal{D}^{test}} \left[ \left( \frac{1}{\tau} \hat{\vw}_{Att}(\mC_{xx}, \mC_{xy}, C_{yy}; \mM, \mV)^T \vx_l + b_{Att}(\vs_x, s_y; \mV) - y_{l} \right)^2  \right],\\
    &\quad = \E_{(\mZ, y_{l}) \sim \mathcal{D}^{test}}  \left[ \left(\frac{1}{\tau} \left(\mM_{11}^T \left( \mC_{xx} \vv_{21} + v_{22} \mC_{xy} \right) \right)^T \vx_l - y_l \right)^2  \right],
\end{align}
where we use the parameters from pretraining \eqref{eq:optimal_parameters} together with Assumptions \ref{assumption:mean_cov_bound} and \ref{assumption:jointly_diverge} to reach the last line. Then,
\begin{align}
    &\mathcal{G}(\mV, \mM) = \E_{(\mZ, y_{l}) \sim \mathcal{D}^{test}}  \left[ \left(\frac{1}{\tau} \left(\mM_{11}^T \left( \mC_{xx} \vv_{21} + v_{22} \mC_{xy} \right) \right)^T \vx_l - y_l \right)^2  \right], \\
    &\quad = \E \left[ \left(\frac{1}{\tau} \left(\mM_{11}^T \left(\frac{1}{l} \sum_{i \leq l} \bar{\vx}_i \bar{\vx}_i^T \vv_{21} + v_{22} \frac{1}{l} \sum_{i \leq l-1} \bar{\vx}_i (\bar{\vx}_i^T\vw + \epsilon_i)  \right) \right)^T \vx_l - \vw^T \vx_l - \epsilon_l \right)^2  \right],\\
    &\quad = \E  \left[ \left(\frac{1}{\tau} \left(\mM_{11}^T \left(\frac{1}{l} \sum_{i \leq l} \bar{\vx}_i \bar{\vx}_i^T \vv_{21} + v_{22} \frac{1}{l} \sum_{i \leq l-1} \bar{\vx}_i (\bar{\vx}_i^T\vw + \epsilon_i)  \right) \right)^T \vx_l - \vw^T \vx_l \right)^2  \right] + \sigma^2
    \label{eq:in_derivation_with_x_l}
\end{align}
where $\bar{\vx}_i := \vx_i - \vs_x = \vx_i - \frac{1}{l} \sum_{i \leq l} \vx_i$ and we use $\epsilon_l \sim \mathcal{N}(0, \sigma^2)$ to reach the final line. We continue by defining
\begin{align}
    \vw_{diff} := \frac{1}{\tau} \mM_{11}^T \left(\frac{1}{l} \sum_{i \leq l-1} \bar{\vx}_i \bar{\vx}_i^T \vv_{21} + v_{22} \frac{1}{l} \sum_{i \leq l-1} \bar{\vx}_i (\bar{\vx}_i^T\vw + \epsilon_i)  \right) - \vw,
\end{align}
which allows us to write
\begin{align}
    \mathcal{G}(\mV, \mM) &= \E \left[ \left( \vw_{diff}^T \vx_l\right)^2  \right] + \sigma^2, \label{eq:in_derivation_without_x_l}\\
    &= \E \left[ \vw_{diff}^T \E_{\vx_l} [\vx_l \vx_l^T] \vw_{diff}   \right] + \sigma^2,\\
    &= \E \left[ \vw_{diff}^T \left(\vmu_x \vmu_x^T + \mSigma_x \right) \vw_{diff}   \right] + \sigma^2,
\end{align}
by the law of total expectation since $\vw_{diff}$ is independent of $\vx_l$. Note that when writing \eqref{eq:in_derivation_without_x_l}, we safely ignore terms with $(1/l)\bar{\vx}_l \bar{\vx}_l^T \vv_{21}$ in \eqref{eq:in_derivation_with_x_l} since they vanish as $l \to \infty$ by Assumptions \ref{assumption:mean_cov_bound}-\ref{assumption:jointly_diverge} and \ref{asssumption:parameters_M_V}. Letting $\mA := \vmu_x \vmu_x^T + \mSigma_x$, we write 
\begin{align}
    \mathcal{G}(\mV, \mM) &= \E \left[ \vw_{diff}^T \mA \vw_{diff}   \right] + \sigma^2,\\
    &= \E \left[ \text{Tr} \left(\vw_{diff}^T \mA \vw_{diff} \right)   \right] + \sigma^2,\\
    &= \E \left[ \text{Tr} \left(\mA \vw_{diff} \vw_{diff}^T \right)   \right] + \sigma^2,\\
    &= \text{Tr}\left(\mA \E[\vw_{diff} \vw_{diff}^T] \right) + \sigma^2,
    \label{eq:in_derivation_generalization_error}
\end{align}
where we first apply the cyclic property of trace and then use the linearity of expectation and trace to reach the last line. Now, we need to calculate $\E[\vw_{diff} \vw_{diff}^T]$, for which we first take the expectation over $\vw$. To do so, we rewrite $\vw_{diff}$ as 
\begin{align}
    \vw_{diff} &= \underbrace{\frac{1}{\tau} \mM_{11}^T \left(\frac{1}{l} \sum_{i \leq l-1} \bar{\vx}_i \bar{\vx}_i^T \vv_{21} + v_{22} \frac{1}{l} \sum_{i \leq l-1} \bar{\vx}_i \epsilon_i  \right)}_{\ve} + \underbrace{\left(\frac{v_{22}}{\tau} \mM_{11}^T \frac{1}{l} \sum_{i \leq l-1} \bar{\vx}_i \bar{\vx}_i^T -\mI \right)}_{\mD}\vw,\\
    &= \ve + \mD\vw,
\end{align}
where we define
\begin{align}
    \ve &:= \frac{1}{\tau} \mM_{11}^T \left(\frac{1}{l} \sum_{i \leq l-1} \bar{\vx}_i \bar{\vx}_i^T \vv_{21} + v_{22} \frac{1}{l} \sum_{i \leq l-1} \bar{\vx}_i \epsilon_i  \right), \\
    \mD &:= \left(\frac{v_{22}}{\tau} \mM_{11}^T \frac{1}{l} \sum_{i \leq l-1} \bar{\vx}_i \bar{\vx}_i^T -\mI \right).
\end{align}
Since $\ve$ and $\mD$ are independent of $\vw$, we can easily calculate $\E_{\vw}[\vw_{diff} \vw_{diff}^T]$ as follows
\begin{align}
    &\E\left[\E_{\vw}[\vw_{diff} \vw_{diff}^T] \right]= \E\left[ \E_{\vw}[(\ve + \mD\vw) (\ve + \mD\vw)^T ] \right],\\
    &\quad = \E\left[\ve \ve^T \right] + \E\left[\ve \vmu_w^T \mD^T \right]  + \E\left[ \mD \vmu_w \ve^T \right] + \E\left[\mD (\vmu_x \vmu_x^T + \mSigma_w) \mD^T \right],\\
    &\quad = \E\left[ \ve \ve^T \right] + \E\left[ \mD \vmu_w \ve^T \right]^T  + \E\left[ \mD \vmu_w \ve^T \right] + \E\left[ \mD \mB \mD^T \right],
    \label{eq:second_moment_of_w_diff}
\end{align}
where we first apply the law of total expectation, then take the expectation over $\vw$ and finally, we define $\mB := \vmu_x \vmu_x^T + \mSigma_w$ to reach the last line. Note that $\vmu_w$ and $\mB$ are fixed while $\ve$ and $\mD$ are random in the last line. Therefore, we are required to calculate the three expectations that appeared in \eqref{eq:second_moment_of_w_diff}. 

Before getting into the calculations of the aforementioned expectations, we provide the following lemma that is useful for the calculation of the expectations.

\begin{lemma}
Let $\bar{\vx} \sim \mathcal{N}(\mathbf{0}, \mSigma)$, where $\bar{\vx} \in \mathbb{R}^d$. Let $\bar{\vx}_i$ be $l-1$ independent samples of $\bar{\vx}$ for $i=1, \ldots, l-1$. Furthermore, let $\mA$ be a fixed $d \times d$ matrix. Then, the following holds
\begin{align}
    \E\left[\left(\frac{1}{l} \sum_{i\leq l-1} \bar{\vx}_i \bar{\vx}_i^T \right) \mA \left(\frac{1}{l} \sum_{i\leq l-1} \bar{\vx}_i \bar{\vx}_i^T \right) \right] = \frac{l-1}{l} \mSigma \mA \mSigma + \frac{1}{l} \mSigma \mA^T \mSigma  + \frac{1}{l} \text{Tr}(\mA\mSigma)\mSigma.
\end{align}
\begin{proof}
    This is proven by using Isserlis' theorem \citep{isserlis1918} in Appendix \ref{appendix:proof_of_lemma_forth_moment}.
\end{proof}
\label{lemma:forth_moment_of_gaussian_vectors}
\end{lemma}
Note that our inputs $\bar{\vx}_i$ are centered, i.e., $\bar{\vx}_i = \vx_i - \frac{1}{l} \sum_{i\leq l} \vx_i$, so their distribution is $\mathcal{N}(\mathbf{0}, \mSigma_x)$ as $l \to \infty$. Therefore, Lemma \ref{lemma:forth_moment_of_gaussian_vectors} is directly applicable in our setting. 

Next, we start the calculations of the expectations in \eqref{eq:second_moment_of_w_diff} with $\E\left[ \ve \ve^T \right]$ as follows
\begin{align}
    &\E\left[ \ve \ve^T \right] = \frac{1}{\tau^2} \mM_{11}^T \E \left[ \left(\frac{1}{l} \sum_{i \leq l-1} \bar{\vx}_i \bar{\vx}_i^T \vv_{21} + v_{22} \frac{1}{l} \sum_{i \leq l-1} \bar{\vx}_i \epsilon_i  \right) \right. \nonumber\\
    &\qquad \qquad \qquad \qquad \qquad \qquad \left. \cdot \left(\frac{1}{l} \sum_{i \leq l-1} \vv_{21}^T\bar{\vx}_i \bar{\vx}_i^T  + v_{22} \frac{1}{l} \sum_{i \leq l-1} \bar{\vx}_i^T \epsilon_i  \right) \right] \mM_{11},\\
    &\; = \frac{1}{\tau^2} \mM_{11}^T \left( \E \left[ \left(\frac{1}{l} \sum_{i \leq l-1} \bar{\vx}_i \bar{\vx}_i^T \vv_{21} \right) \left(\frac{1}{l} \sum_{i \leq l-1} \vv_{21}^T \bar{\vx}_i \bar{\vx}_i^T  \right) \right. \right. \nonumber\\
    &\qquad \qquad \qquad \qquad \qquad \qquad \left. \left. + \left(v_{22} \frac{1}{l} \sum_{i \leq l-1} \bar{\vx}_i \epsilon_i  \right) \left(v_{22} \frac{1}{l} \sum_{i \leq l-1} \bar{\vx}_i^T \epsilon_i  \right) \right] \right) \mM_{11},\\
    &\; = \frac{1}{\tau^2} \mM_{11}^T \left( \E \left[ \left(\frac{1}{l} \sum_{i \leq l-1} \bar{\vx}_i \bar{\vx}_i^T\right) \vv_{21}\vv_{21}^T \left(\frac{1}{l} \sum_{i \leq l-1}  \bar{\vx}_i \bar{\vx}_i^T  \right) + \left(v_{22}^2 \frac{\sigma^2}{l^2} \sum_{i \leq l-1} \bar{\vx}_i\bar{\vx}_i^T  \right) \right] \right) \mM_{11},\\
    &\; = \frac{1}{\tau^2} \mM_{11}^T \left( \mSigma_x \vv_{21} \vv_{21}^T \mSigma_x + \frac{1}{l} \text{Tr} \left(\vv_{21} \vv_{21}^T \mSigma_x \right)  \mSigma_x  +  v_{22}^2 \frac{\sigma^2 (l-1)}{l^2} \mSigma_x \right) \mM_{11},\\
    &\; = \frac{1}{\tau^2} \mM_{11}^T \left( v_{22}^2 \frac{\sigma^2}{l} \mSigma_x \right) \mM_{11},
\end{align}
where we first use the independence of the random variables and $\epsilon_i \sim \mathcal{N}(0,\sigma^2)$ to simplify the equation. Then, we apply Lemma \ref{lemma:forth_moment_of_gaussian_vectors} and use the fact that $\E[\bar{\vx}_i\bar{\vx}_i^T] = \mSigma_x$ to get the penultimate line. Finally, we drop the vanishing terms and simplify the result using Assumptions \ref{assumption:mean_cov_bound}-\ref{assumption:jointly_diverge} and \ref{asssumption:parameters_M_V} in order to reach the last line.

We continue with the calculation of $\E\left[ \mD \vmu_w \ve^T \right]$ as
\begin{align}
    &\E\left[ \mD \vmu_w \ve^T \right] \nonumber\\
    &\; = \frac{1}{\tau} \E \left[ \left(\frac{v_{22}}{\tau} \mM_{11}^T \frac{1}{l} \sum_{i \leq l-1} \bar{\vx}_i \bar{\vx}_i^T -\mI \right) \vmu_w \left(\frac{1}{l} \sum_{i \leq l-1} \vv_{21}^T \bar{\vx}_i \bar{\vx}_i^T  + v_{22} \frac{1}{l} \sum_{i \leq l-1} \bar{\vx}_i^T \epsilon_i  \right) \right] \mM_{11},\\
    &\; = \frac{1}{\tau} \E \left[ \left(\frac{v_{22}}{\tau} \mM_{11}^T \frac{1}{l} \sum_{i \leq l-1} \bar{\vx}_i \bar{\vx}_i^T -\mI \right) \vmu_w \left(\frac{1}{l} \sum_{i \leq l-1} \vv_{21}^T \bar{\vx}_i \bar{\vx}_i^T \right) \right] \mM_{11},\\
    &\; = \frac{1}{\tau} \frac{v_{22}}{\tau} \mM_{11}^T \E \left[\left( \frac{1}{l} \sum_{i \leq l-1} \bar{\vx}_i \bar{\vx}_i^T \right) \vmu_w \vv_{21}^T \left(\frac{1}{l} \sum_{i \leq l-1} \bar{\vx}_i \bar{\vx}_i^T \right)\right]\mM_{11} \nonumber\\
    &\qquad - \vmu_w \vv_{21}^T \E \left[\frac{1}{l} \sum_{i \leq l-1} \bar{\vx}_i \bar{\vx}_i^T \right] \mM_{11},\\
    &\; = \frac{1}{\tau} \frac{v_{22}}{\tau} \mM_{11}^T \left( \mSigma_x \vmu_w \vv_{21}^T \mSigma_x + \frac{1}{l}  \mSigma_x \vv_{21} \vmu_w^T \mSigma_x + \frac{1}{l} \text{Tr}\left( \vmu_w \vv_{21}^T \mSigma_x \right) \mSigma_x  \right) \mM_{11} \nonumber\\
    &\qquad -  \frac{1}{\tau} \frac{l-1}{l} \vmu_w \vv_{21}^T \mSigma_x \mM_{11},\\
    &\; = \frac{v_{22}}{\tau^2} \mM_{11}^T \left( \mSigma_x \vmu_w \vv_{21}^T \mSigma_x  + \frac{1}{l}\text{Tr}\left( \vmu_w \vv_{21}^T \mSigma_x \right) \mSigma_x  \right) \mM_{11} -  \frac{1}{\tau}  \vmu_w \vv_{21}^T \mSigma_x \mM_{11},
\end{align}
where we again first use the independence of the random variables and $\epsilon_i \sim \mathcal{N}(0,\sigma^2)$. Then, we apply basic algebraic manipulations. To reach the penultimate line, we utilize Lemma \ref{lemma:forth_moment_of_gaussian_vectors} together with the fact that $\E[\bar{\vx}_i\bar{\vx}_i^T] = \mSigma_x$. Using Assumptions \ref{assumption:mean_cov_bound}-\ref{assumption:jointly_diverge} and \ref{asssumption:parameters_M_V}, we reach the last line.

Finally, we calculate $\E\left[ \mD \mB \mD^T \right]$ as follows
\begin{align}
    &\E\left[ \mD \mB \mD^T \right] \nonumber\\
    &\; = \E \left[ \left(\frac{v_{22}}{\tau} \mM_{11}^T \frac{1}{l} \sum_{i \leq l-1} \bar{\vx}_i \bar{\vx}_i^T -\mI \right) \mB \left(\frac{v_{22}}{\tau}  \frac{1}{l} \sum_{i \leq l-1} \bar{\vx}_i \bar{\vx}_i^T \mM_{11} -\mI \right) \right],\\
    &\; = \E \left[ \left(\frac{v_{22}}{\tau} \mM_{11}^T \frac{1}{l} \sum_{i \leq l-1} \bar{\vx}_i \bar{\vx}_i^T \right) \mB \left(\frac{v_{22}}{\tau}  \frac{1}{l} \sum_{i \leq l-1} \bar{\vx}_i \bar{\vx}_i^T \mM_{11} \right) \right] \nonumber \\
    & \qquad - \E \left[\left(\frac{v_{22}}{\tau} \mM_{11}^T \frac{1}{l} \sum_{i \leq l-1} \bar{\vx}_i \bar{\vx}_i^T \right) \mB \right] - \E\left[ \mB \left(\frac{v_{22}}{\tau}  \frac{1}{l} \sum_{i \leq l-1} \bar{\vx}_i \bar{\vx}_i^T \mM_{11} \right) \right] + \mB,\\
    &\; = \frac{v_{22}^2}{\tau^2} \mM_{11}^T\E \left[ \left( \frac{1}{l} \sum_{i \leq l-1} \bar{\vx}_i \bar{\vx}_i^T \right) \mB \left( \frac{1}{l} \sum_{i \leq l-1} \bar{\vx}_i \bar{\vx}_i^T  \right) \right] \mM_{11} \nonumber \\
    & \qquad - \frac{v_{22}}{\tau} \mM_{11}^T \E \left[\left(  \frac{1}{l} \sum_{i \leq l-1} \bar{\vx}_i \bar{\vx}_i^T \right) \right] \mB - \frac{v_{22}}{\tau}  \mB \E\left[  \left( \frac{1}{l} \sum_{i \leq l-1} \bar{\vx}_i \bar{\vx}_i^T  \right) \right] \mM_{11} + \mB,\\
    &\; = \frac{v_{22}^2}{\tau^2} \mM_{11}^T \left( \mSigma_x \mB \mSigma_x + \frac{1}{l} \text{Tr} \left( \mB \mSigma_x \right) \mSigma_x \right) \mM_{11} - \frac{v_{22}}{\tau} \frac{l-1}{l} \mM_{11}^T \mSigma_x \mB \nonumber\\
    &\qquad - \frac{v_{22}}{\tau} \frac{l-1}{l} \mB \mSigma_x \mM_{11} + \mB,\\
    &\; = \frac{v_{22}^2}{\tau^2} \mM_{11}^T \left( \mSigma_x \mB \mSigma_x + \frac{1}{l} \text{Tr} \left( \mB \mSigma_x \right) \mSigma_x \right) \mM_{11} - \frac{v_{22}}{\tau}  \mM_{11}^T \mSigma_x \mB - \frac{v_{22}}{\tau}\mB \mSigma_x \mM_{11} + \mB,
\end{align}
where we first do basic algebraic manipulations. Then, we use Lemma \ref{lemma:forth_moment_of_gaussian_vectors} and $\E[\bar{\vx}_i\bar{\vx}_i^T] = \mSigma_x$ to get the penultimate line. For the final line, we utilize $l \to \infty$ by Assumption \ref{assumption:jointly_diverge}.

Putting the found expectation results into \eqref{eq:second_moment_of_w_diff}, we get
\begin{align}
    \E\left[\E_{\vw}[\vw_{diff} \vw_{diff}^T] \right]&= \E\left[ \ve \ve^T \right] + \E\left[ \mD \vmu_w \ve^T \right]^T  + \E\left[ \mD \vmu_w \ve^T \right] + \E\left[ \mD \mB \mD^T \right],\\
    &= \frac{1}{\tau^2}\mM_{11}^T \mF_1 \mM_{11} -  \frac{1}{\tau} \mF_2 \mM_{11} + \frac{1}{\tau} \mM_{11}^T \mF_2^T + \mB.
\end{align}
where matrices $\mF_1$ and $\mF_2$ are defined as
\begin{align}
    \mF_1 &:= v_{22}^2 \frac{\sigma^2}{l} \mSigma_x + v_{22} \left( \mSigma_x \vmu_w \vv_{21}^T \mSigma_x + \frac{1}{l}\text{Tr}\left( \vmu_w \vv_{21}^T \mSigma_x \right) \mSigma_x \right)\\
    &\qquad   + v_{22} \left( \mSigma_x \vmu_w \vv_{21}^T \mSigma_x + \frac{1}{l}\text{Tr}\left( \vmu_w \vv_{21}^T \mSigma_x \right) \mSigma_x \right)^T + v_{22}^2 \left( \mSigma_x \mB \mSigma_x + \frac{1}{l} \text{Tr} \left( \mB \mSigma_x \right) \mSigma_x \right), \nonumber\\
    &=  \left( \mSigma_x \hat{\mB}  + \left( v_{22}^2 \frac{\sigma^2}{l} + \frac{1}{l} \text{Tr} \left( \hat{\mB} \mSigma_x \right)\right) \mI \right) \mSigma_x,\\
    \mF_2 &:= \vmu_w \vv_{21}^T \mSigma_x + v_{22} \mB \mSigma_x  = (\vmu_w \vv_{21}^T + v_{22} \mB) \mSigma_x,
\end{align}
with $\hat{\mB} := v_{22} \vmu_w \vv_{21}^T + v_{22} \vv_{21} \vmu_w^T + v_{22}^2 \mB$.

Going back to generalization error in \eqref{eq:in_derivation_generalization_error}, we have
\begin{align}
    \mathcal{G}(\mV, \mM) &= \text{Tr}\left(\mA \E[\vw_{diff} \vw_{diff}^T] \right) + \sigma^2,\\
    &= \text{Tr}\left(\mA \left(\frac{1}{\tau^2}\mM_{11}^T \mF_1 \mM_{11} - \frac{1}{\tau} \mF_2 \mM_{11} + \frac{1}{\tau} \mM_{11}^T \mF_2^T + \mB \right) \right) + \sigma^2,
\end{align}
where $\mF_1 = \left( \mSigma_x \hat{\mB}  + \frac{1}{l} \left( v_{22}^2\sigma^2 + \text{Tr} \left( \hat{\mB} \mSigma_x \right)\right) \mI \right) \mSigma_x$, and $\mF_2 = (\vmu_w \vv_{21}^T + v_{22} \mB) \mSigma_x$. Furthermore, $\hat{\mB}$ is defined as  $\hat{\mB} := v_{22} \vmu_w \vv_{21}^T + v_{22} \vv_{21} \vmu_w^T + v_{22}^2 \mB$.

\section{Proof of Lemma \ref{lemma:forth_moment_of_gaussian_vectors}}
\label{appendix:proof_of_lemma_forth_moment}
We first restate the lemma as follows.

Let $\bar{\vx} \sim \mathcal{N}(0, \mSigma)$, where $\bar{\vx} \in \mathbb{R}^d$. Let $\bar{\vx}_i$ be $l$ independent samples of $\bar{\vx}$ for $i=1, \ldots, l$. Let $\mA$ be a fixed $d \times d$ matrix. Then, the following holds
\begin{align}
    \E\left[\left(\frac{1}{l} \sum_{i\leq l} \bar{\vx}_i \bar{\vx}_i^T \right) \mA \left(\frac{1}{l} \sum_{i\leq l} \bar{\vx}_i \bar{\vx}_i^T \right) \right] = \mSigma \mA \mSigma + \frac{1}{l} \mSigma \mA^T \mSigma + \frac{1}{l} \text{Tr}(\mA \mSigma) \mSigma.
\end{align}
\begin{proof}
    Let $\mS_x = \frac{1}{l} \sum_{i=1}^l \bar{\vx}_i \bar{\vx}_i^T$. First, note that $E[\bar{\vx}_i \bar{\vx}_i^T] = \mSigma$ since $\bar{\vx}_i \sim \mathcal{N}(0, \mSigma)$. 
    
    Thus, $\E[\mS_x] = \frac{1}{l} \sum_{i=1}^l \E[\bar{\vx}_i \bar{\vx}_i^T] = \frac{1}{l} \sum_{i=1}^l \mSigma = \mSigma$. We have
\begin{align}
    \mS_x \mA \mS_x = \frac{1}{l^2} \sum_{i=1}^l \sum_{j=1}^l \bar{\vx}_i \bar{\vx}_i^T \mA \bar{\vx}_j \bar{\vx}_j^T
\end{align}
Taking the expectation, we get
\begin{align}
    \E[\mS_x \mA \mS_x] = \frac{1}{l^2} \sum_{i=1}^l \sum_{j=1}^l \E[\bar{\vx}_i \bar{\vx}_i^T \mA \bar{\vx}_j \bar{\vx}_j^T]
\end{align}

When $i \neq j$, $\bar{\vx}_i$ and $\bar{\vx}_j$ are independent, so
\begin{align}
    \E[\bar{\vx}_i \bar{\vx}_i^T \mA \bar{\vx}_j \bar{\vx}_j^T] = \E[\bar{\vx}_i \bar{\vx}_i^T] \mA \E[\bar{\vx}_j \bar{\vx}_j^T] = \mSigma \mA \mSigma
\end{align}
When $i=j$,
\begin{align}
    \E[\bar{\vx}_i \bar{\vx}_i^T \mA \bar{\vx}_i \bar{\vx}_i^T] = \E[\bar{\vx} \bar{\vx}^T \mA \bar{\vx} \bar{\vx}^T]
\end{align}
Let $\bar{\vx} = [x_1, x_2, \dots, x_d]^T$. Then, from Isserlis' theorem \citep{isserlis1918}, we have
\begin{align}
    \E[x_i x_j x_k x_l] = \mSigma_{ij} \mSigma_{kl} + \mSigma_{ik} \mSigma_{jl} + \mSigma_{il} \mSigma_{jk}
\end{align}
Let $\mA = [a_{ij}]$. Then, $\bar{\vx}^T \mA \bar{\vx} = \sum_{i,j} a_{ij} x_i x_j$. Thus, we reach
\begin{align}
    \bar{\vx} \bar{\vx}^T \mA \bar{\vx} \bar{\vx}^T &= \bar{\vx} \bar{\vx}^T \sum_{i,j} a_{ij} x_i x_j,\\
    \E[\bar{\vx}_i \bar{\vx}_i^T \mA \bar{\vx}_i \bar{\vx}_i^T] &=  \text{Tr}(\mA \mSigma) \mSigma + \mSigma \mA \mSigma + \mSigma \mA^T \mSigma.
\end{align}
There are $l^2$ terms in the double sum. $l$ terms are of the form $\E[\bar{\vx}_i \bar{\vx}_i^T \mA \bar{\vx}_i \bar{\vx}_i^T]$ and $l^2 - l$ terms are of the form $\mSigma \mA \mSigma$. Therefore, we can write
\begin{align}
    \E[\mS_x \mA \mS_x] &= \frac{1}{l^2} [l (\text{Tr}(\mA \mSigma) \mSigma + \mSigma \mA \mSigma + \mSigma \mA^T \mSigma) + l(l-1) \mSigma \mA \mSigma],\\
    &= \frac{1}{l} (\text{Tr}(\mA \mSigma) \mSigma + \mSigma \mA \mSigma + \mSigma \mA^T \mSigma) + \frac{l-1}{l} \mSigma \mA \mSigma,\\
    &= \mSigma \mA \mSigma + \frac{1}{l} \mSigma \mA^T \mSigma + \frac{1}{l} \text{Tr}(\mA \mSigma) \mSigma,
\end{align}
which completes the proof.
\end{proof}

\section{Analysis of optimal temperature for ICL under distribution shift}
\label{appendix:optimal_temperature}
Here, we find the optimal temperature minimizing the generalization error. First, recall that we have the following generalization error.
\begin{align}
     \mathcal{G}(\mV, \mM) &= \frac{1}{\tau^2} \text{Tr}\left(\mA \mM_{11}^T \mF_1 \mM_{11} \right)  -  \frac{1}{\tau} \text{Tr}\left(\mA \left(\mF_2 \mM_{11} + \mM_{11}^T \mF_2^T \right)\right)  + \text{Tr}\left(\mA \mB \right) + \sigma^2,
\end{align}
as specified in Theorem \ref{theorem:generalization_error}. So, we can express the generalization error as,
\begin{align}
     \mathcal{G}(\tau; \mV, \mM) &= \frac{a}{\tau^2}  -  \frac{b}{\tau} + c,
\end{align}
where $a := \text{Tr}\left(\mA \mM_{11}^T \mF_1 \mM_{11} \right)$, $b := \text{Tr}\left(\mA \left(\mF_2 \mM_{11} + \mM_{11}^T \mF_2^T \right)\right)$, and $c = \text{Tr}\left(\mA \mB \right) + \sigma^2$. Therefore, we have the following optimization problem
\begin{align}
    \tau_{\text{opt}} &:= \argmin_{\tau} \mathcal{G}(\tau; \mV, \mM),\\
    &=  \argmin_{\tau} \left\{ \frac{a}{\tau^2}  -  \frac{b}{\tau} + c \right\}.
\end{align}

To find the optimal value of $\tau$ that minimizes the given function, we can take the derivative of the expression with respect to $\tau$ and set it to zero. From now on, we consider generalization error as a function of $\tau$, written as $\mathcal{G}(\tau)$. 

Next, find the derivative of $\mathcal{G}(\tau)$ with respect to $\tau$ as
\begin{align}
    \mathcal{G}'(\tau) = -2a\tau^{-3} + b\tau^{-2}.
\end{align}

To find the critical points, set $\mathcal{G}'(\tau) = 0$ as follows
\begin{align}
    \mathcal{G}'(\tau) = -2a\tau^{-3} + b\tau^{-2} = 0,
\end{align}
Solving this equation for $\tau$, we reach the following critical point
\begin{align}
    \tau = \frac{2a}{b}.
\end{align}

Now, we need to check if this is a minimum by taking the second derivative, which is
\begin{align}
    \mathcal{G}''(\tau) = 6a\tau^{-4} - 2b\tau^{-3}.
\end{align}

Evaluate $\mathcal{G}''(\tau)$ at $\tau = \frac{2a}{b}$ as follows
\begin{align}
    \mathcal{G}''\left(\frac{2a}{b}\right) = 6a\left(\frac{2a}{b}\right)^{-4} - 2b\left(\frac{2a}{b}\right)^{-3} = 6a\left(\frac{b^4}{16a^4}\right) - 2b\left(\frac{b^3}{8a^3}\right) = \frac{b^4}{8a^3}.
\end{align}
Since $a, b > 0$, we reach $\mathcal{G}''\left(\frac{2a}{b}\right) = \frac{b^4}{8a^3} > 0$, which means the function has a minimum at $\tau = \frac{2a}{b}$. Therefore, $\tau_{\text{opt}} = \frac{2a}{b}$ is the solution minimizing the generalization error $\mathcal{G}(\tau)$. Writing $a,b$ back into the optimal solution, we get
\begin{align}
    \tau_{\text{opt}} = \frac{2 \text{Tr}\left(\mA \mM_{11}^T \mF_1 \mM_{11} \right)}{\text{Tr}\left(\mA \left(\mF_2 \mM_{11} + \mM_{11}^T \mF_2^T \right)\right)},
\end{align}
which concludes our derivation of the optimal temperature $\tau_{\text{opt}}$.

\section{An insight driven from optimal temperature for other settings}
\label{appendix:insights_for_other_settings}

In this section, we extract a mathematical heuristic from the optimal temperature in Theorem~\ref{theorem:optimal_temperature} that can be applied to ICL settings beyond our existing setting involving approximate softmax attention and regression tasks.  
Specifically, we consider Transformers employing standard softmax attention.  
Recall that the attention temperature scales the pre-softmax scores (i.e., $(\mK \mZ)^{\top} (\mQ \mZ)$ in~\eqref{eq:softmax_attention}), thereby controlling the variance of the final scores.  
Since the optimal temperature depends on the distribution of these scores, it can be naturally characterized by the moments of that distribution.  
Our central intuition is that the optimal temperature identified in Theorem~\ref{theorem:optimal_temperature} relates directly to the first two moments of the pre-softmax scores.  
Although this optimal temperature was derived for \emph{approximate softmax} attention, the insight remains relevant for softmax attention because the two mechanisms behave similarly in the regime considered (see Appendix~\ref{appendix:temperature_in_approximate_softmax}).

We now illustrate how the optimal temperature in Theorem~\ref{theorem:optimal_temperature} can be related to the first two moments of the pre-softmax scores.  
For simplicity, we consider the case $\vmu_x=\vmu_w=\vm_{21}=\mathbf{0}$ and $\mSigma_w=\mI$, under which the optimal temperature reduces to
\begin{align}
    \tau_{\text{opt}}
    = 
    \frac{v_{22} 
    \operatorname{Tr}\!\left( 
        \mSigma_x \mM_{11} \mSigma_x \mM_{11}^{\top} \mSigma_x 
    \right)}
    {\frac{1}{2} 
    \operatorname{Tr}\!\left( 
        \mSigma_x 
        \left( \mSigma_x \mM_{11} + \mM_{11}^{\top} \mSigma_x^{\top} \right) 
    \right)}.
    \label{eq:insight_optimal_temperature}
\end{align}
We next show how this expression connects to the first two moments of $(\mK \mZ)^{\top} (\mQ \mZ)$.  
Let $\vz_i$ denote the $i$-th column of $\mZ$ from~\eqref{eq:embedding_matrix} and recall $\mK^{\top}\mQ=\mM$.  
We therefore compute $\E[\vz_i^{\top}\mM\vz_j]$ and $\E[(\vz_i^{\top}\mM\vz_j)^2]$ for $i,j\in\{1,\dots,l\}$.  
Starting with the first moment for $i=j$:
\begin{align}
    \E[\vz_i^{\top}\mM\vz_i]
    &=\operatorname{Tr}\!\big(\E[\vz_i\vz_i^{\top}]\mM\big)
    =\operatorname{Tr}\!\big(\mSigma_x\mM_{11}\big),
\end{align}
where the block structure (and zero entries) of $\mM$ is used in the last step.  
For $i\neq j$,
\begin{align}
    \E[\vz_i^{\top}\mM\vz_j]
    &=\operatorname{Tr}\!\big(\mM\E[\vz_i\vz_j^{\top}]\big)=0,
\end{align}
by independence of $\vz_i$ and $\vz_j$.  
For the second moment with $i\neq j$:
\begin{align}
    \E\!\left[(\vz_i^{\top}\mM\vz_j)^2\right]
    &=\E\!\left[\vz_i^{\top}\mM\vz_j\vz_j^{\top}\mM^{\top}\vz_i\right]\\
    &=\E\!\left[\vx_i^{\top}\mM_{11}\vx_j\vx_j^{\top}\mM_{11}^{\top}\vx_i\right]\\
    &=\E_{\vx_i}\!\left[\vx_i^{\top}\mM_{11}\E_{\vx_j}[\vx_j\vx_j^{\top}]
        \mM_{11}^{\top}\vx_i\right]\\
    &=\E_{\vx_i}\!\left[\vx_i^{\top}\mM_{11}\mSigma_x\mM_{11}^{\top}\vx_i\right]\\
    &=\operatorname{Tr}\!\big(
        \mM_{11}\mSigma_x\mM_{11}^{\top}\E_{\vx_i}[\vx_i\vx_i^{\top}]
    \big)\\
    &=\operatorname{Tr}\!\big(\mM_{11}\mSigma_x\mM_{11}^{\top}\mSigma_x\big),
\end{align}
where we again exploit the block structure of $\mM$ and apply straightforward manipulations.

We observe a parallel between the numerator of~\eqref{eq:insight_optimal_temperature} and the computed second moment (for $i\neq j$), and between the denominator and the first moment (for $i=j$).  
This motivates the heuristic that the optimal temperature should be roughly proportional to the ratio of the second moment (for $i\neq j$) to the first moment (for $i=j$).  
Accordingly, in our LLM experiments (Figure~\ref{fig:LLM}), we select the temperature proportional to this ratio while taking care to avoid numerical issues.

Finally, we note an important caveat: in order to obtain an insight of practical relevance, we intentionally relaxed the rigor applied in our main theoretical results.  
Consequently, the heuristic derived here—and the accompanying empirical findings—should be viewed as preliminary, intended to inspire future work on principled selection of attention temperature in practice.

\section{Experimental details and GPT-2 experiments}
\label{appendix:experimental_details}
This section describes our experimental setups for GPT-2 and large language models (LLMs), including the motivation for our distribution-shift scenarios.

\subsection{GPT-2: Transformer with MLP layers}
\begin{figure}[H]
    \centering
    \includegraphics[width=0.4\linewidth]{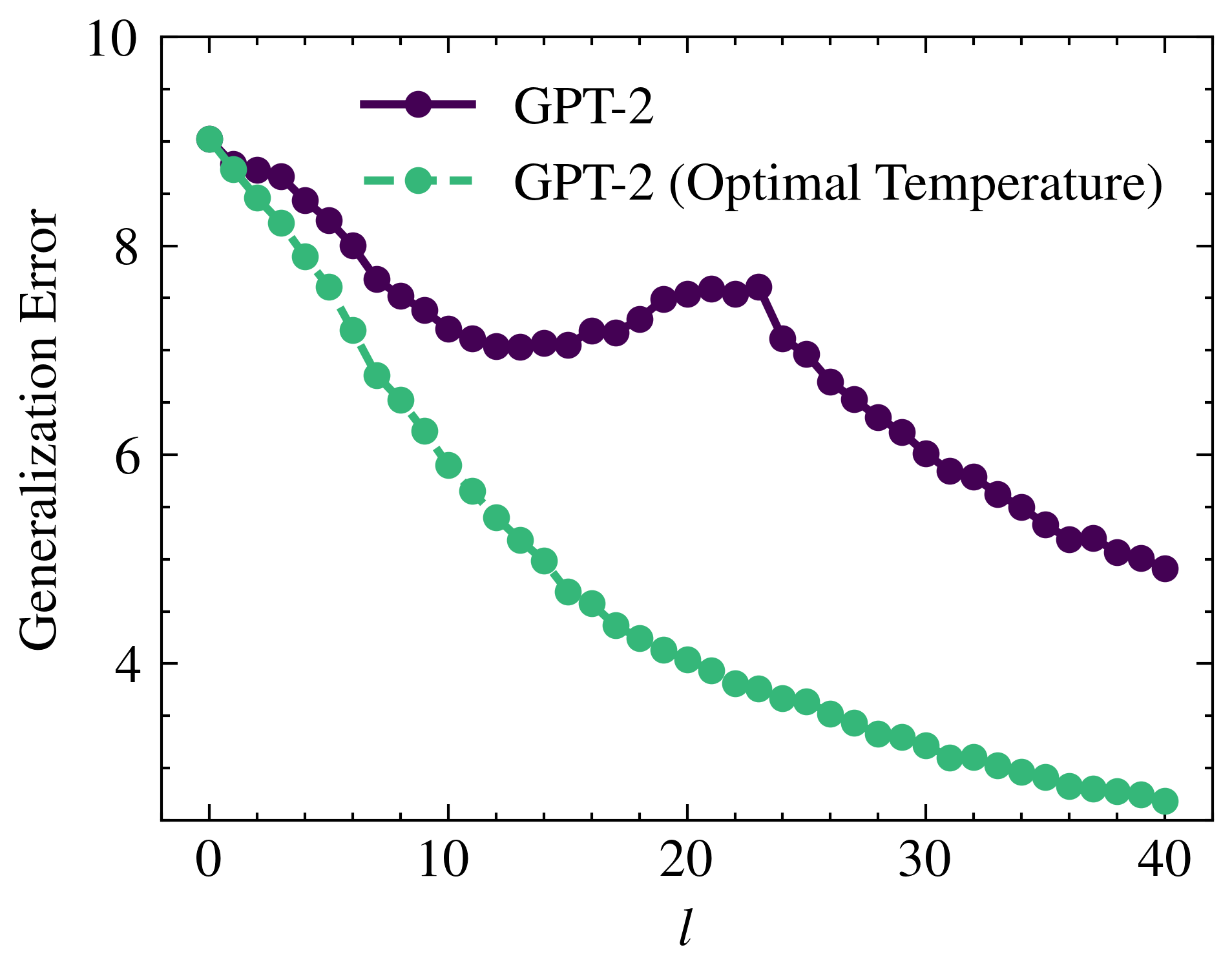}
    \caption{GPT-2~\citep{radford2019language} under an input-covariance shift.
    GPT-2 exemplifies the Transformer architecture~\citep{vaswani2017attention}, combining multi-layer perceptrons with multi-head softmax self-attention.  
    The model here is pretrained by~\citet{garg2022can} on the linear regression tasks defined in~\eqref{eq:data_model}.  
    We consider a shift from $\mSigma_x^{\text{train}}=\mI$ to $\mSigma_x^{\text{test}}=3\mI$.  
    The attention temperature at each layer is scaled as $\tau\sqrt{d_k}$ (where $d_k$ is the key dimension) to ensure dimension-independent $\tau$ values.}
    \label{fig:gpt2_experiment}
\end{figure}

Building on the approximate-attention experiments, we investigate whether the optimal temperature also benefits more complex Transformer models on linear regression tasks.  
We evaluate GPT-2~\citep{radford2019language} under a shift in input covariance (Figure~\ref{fig:gpt2_experiment}).  
Consistent with prior work~\citep{garg2022can, zhang2024trained}, such shifts substantially degrade performance and can even induce nonmonotonic generalization error with respect to context length $l$.  
Remarkably, applying the optimal temperature mitigates this nonmonotonicity and improves in-context generalization.

\subsection{Details of the GPT-2 experiments in Figure \ref{fig:gpt2_experiment}}
We use the standard GPT-2 architecture~\citep{radford2019language} as implemented in HuggingFace~\citep{wolf-etal-2020-transformers}, leveraging the pretrained model of~\citet{garg2022can}. Training data match ours, while their training procedure differs slightly: the loss is auto-regressive, i.e., the average over the entire context sequence of length $l=40$.  We adopt the same embedding method as in~\citet{garg2022can}. The input dimension is $d=20$, with 12 layers and 8 heads.  
All GPT-2 experiments run on an NVIDIA Tesla V100 GPU and complete in approximately 10 minutes.

\subsection{Details of the LLM experiments in Figure \ref{fig:LLM}}
For our large language model experiments, we employ Llama2-7B~\citep{touvron2023llama} and the SCIQ dataset~\citep{welbl2017crowdsourcing}, which contains science questions with supporting information.  
We generate ICL problems following~\citet{gao2024on}, selecting in-context demonstrations using the TopK retrieval technique~\citep{liu2022makes} to ensure relevance.  
An example ICL sample from SCIQ appears in Table~\ref{table:sciq_icl_example}.  
To simulate distribution shift, we follow~\citet{gao2024on} and introduce noisy labels—incorrect but semantically related—to the in-context demonstrations (Appendix~\ref{appendix:why_noisy_labels}).  
Table~\ref{table:sciq_noisy_labels} gives an example.  
The noisy ratio denotes the fraction of demonstrations with noisy labels (e.g., 0.6 means 60\% noisy).  
We modify and use the codebase of~\citet{gao2024on}, built on HuggingFace~\citep{wolf-etal-2020-transformers} and OpenICL~\citep{wu2023openicl}.  
All LLM experiments run on an NVIDIA A40 GPU; a single Monte Carlo run per plot in Figure~\ref{fig:LLM} takes a few hours.

\subsection{Why in-context demonstrations with noisy labels as an example of distribution shift?}
\label{appendix:why_noisy_labels}
The link between noisy labels in demonstrations and distribution shift may not be immediately obvious.  
Quantifying pretraining–test shifts for pretrained LLMs is inherently difficult because their pretraining data are complex mixtures of sources~\citep{touvron2023llama}.  
However, we hypothesize—following~\citet{gao2024on}—that high perplexity can serve as an empirical indicator of distribution shift.  
Inputs aligned with the training distribution tend to yield low perplexity (high-confidence generation), whereas contradictory or out-of-distribution inputs induce high perplexity.  
Since noisy demonstrations are expected to contradict training-set patterns, they yield high perplexity and thereby act as a proxy for distribution shift.  
Consequently, introducing noisy labels into in-context demonstrations constitutes a principled way to test the robustness of in-context learning under distribution shift.

\begin{table}[H]
  \centering
  \setlength{\tabcolsep}{4pt}
  \begin{tabularx}{\textwidth}{@{}p{0.18\textwidth}X@{}}
    \toprule
    \multicolumn{2}{@{}l@{}}{\textbf{In-context demonstration 1}} \\
    \midrule
    \textbf{Support:} & Cells are organized into tissues, tissues are organized into organs. \\
    \textbf{Question:} & What is considered the smallest unit of the organ? \\
    \textbf{Answer:} & \textcolor{blue}{Cells} \\
    \midrule
    \multicolumn{2}{@{}l@{}}{\textbf{In-context demonstration 2}} \\
    \midrule
    \textbf{Support:} & \dots four basic types of tissue: connective, muscle, \textcolor{blue}{nervous}, and epithelial. \\
    \textbf{Question:} & The four basic types of tissue are epithelial, muscle, connective, and what? \\
    \textbf{Answer:} & \textcolor{blue}{nervous} \\
    \midrule
    \multicolumn{2}{@{}l@{}}{\vdots} \\
    \midrule
    \multicolumn{2}{@{}l@{}}{\textbf{Test example}} \\
    \midrule
    \textbf{Support:} & All forms of life are built of at least one cell. A cell is the basic unit of life. \\
    \textbf{Question:} & What are the smallest structural and functional units of all living organisms? \\
    \textbf{Output:} & ??? \\
    \bottomrule
    \vspace{0.5em}
  \end{tabularx}
  \caption{A sample illustration of in-context learning on the SCIQ dataset.}
  \label{table:sciq_icl_example}
\end{table}

\begin{table}[H]
    \centering
    \begin{tabular}{l p{11cm}}
    \toprule
    \textbf{Setting} & \textbf{In-context demonstration} \\
    \midrule
    \textcolor{blue}{True Label} & 
    \textbf{Support:} \textcolor{blue}{Cells} are organized into tissues, tissues are organized into organs. \\
    & \textbf{Question:} What is considered the smallest unit of the organ? \\
    & \textbf{Label:} \textcolor{blue}{Cells} \\
    \addlinespace[0.5em]
    \textcolor{red}{Noisy Label} &
    \textbf{Support:} Cells are organized into tissues, \textcolor{red}{tissues} are organized into organs. \\
    & \textbf{Question:} What is considered the smallest unit of the organ? \\
    & \textbf{Label:} \textcolor{red}{tissues} \\
    \bottomrule
    \vspace{0.2em}
    \end{tabular}
    \caption{An example of a true label vs. a relevant but noisy label. A relevant label is related to the question but is not necessarily true. Therefore, relevant labels can be considered noisy labels.}
    \label{table:sciq_noisy_labels}
\end{table}

\section{Numerical illustration of the optimal temperature and its generalization behavior in the setting of Section \ref{section:interpretation_optimal_temperature}}

To complement the analytical reductions in the setting of Section \ref{section:interpretation_optimal_temperature}, we now present numerical experiments illustrating how the optimal attention temperature varies under different types of distribution shift. These simulations closely follow the structure predicted by the closed-form expression in Theorem~\ref{theorem:optimal_temperature} and its simplified forms in~\eqref{eq:derived_example_optimal_temperature} and~\eqref{eq:moment_ratio_with_correction}.

\begin{figure*}[h!]
    \centering
    \begin{subfigure}[b]{0.32\textwidth}
         \centering
         \includegraphics[width=0.99\linewidth]{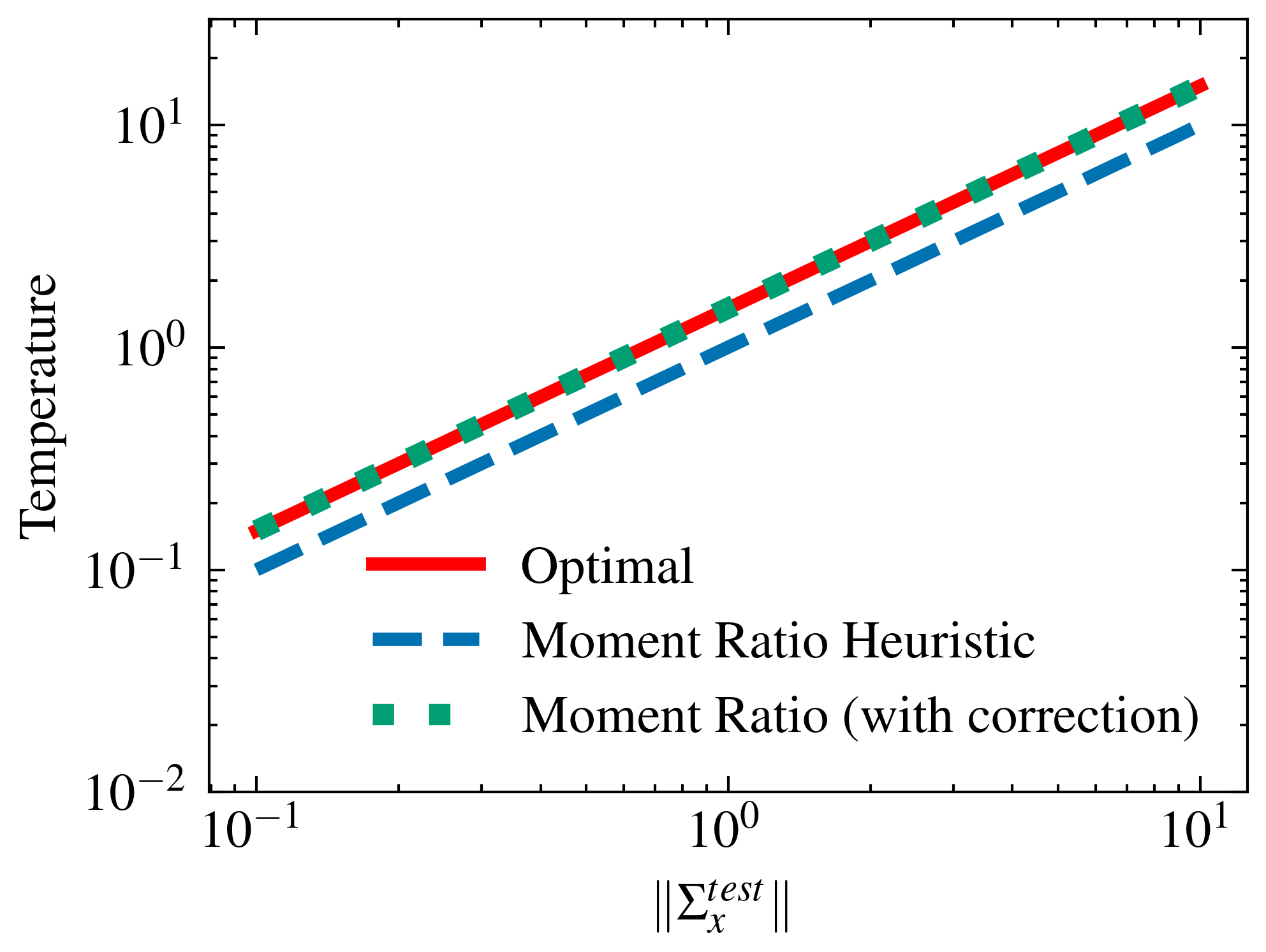}
         \captionsetup{justification=centering}
         \caption{Input Covariance Shifted\\ ($\mSigma_x^{test}= a \mSigma_x^{train})$}
     \end{subfigure}
     \begin{subfigure}[b]{0.32\textwidth}
         \centering
         \includegraphics[width=0.99\linewidth]{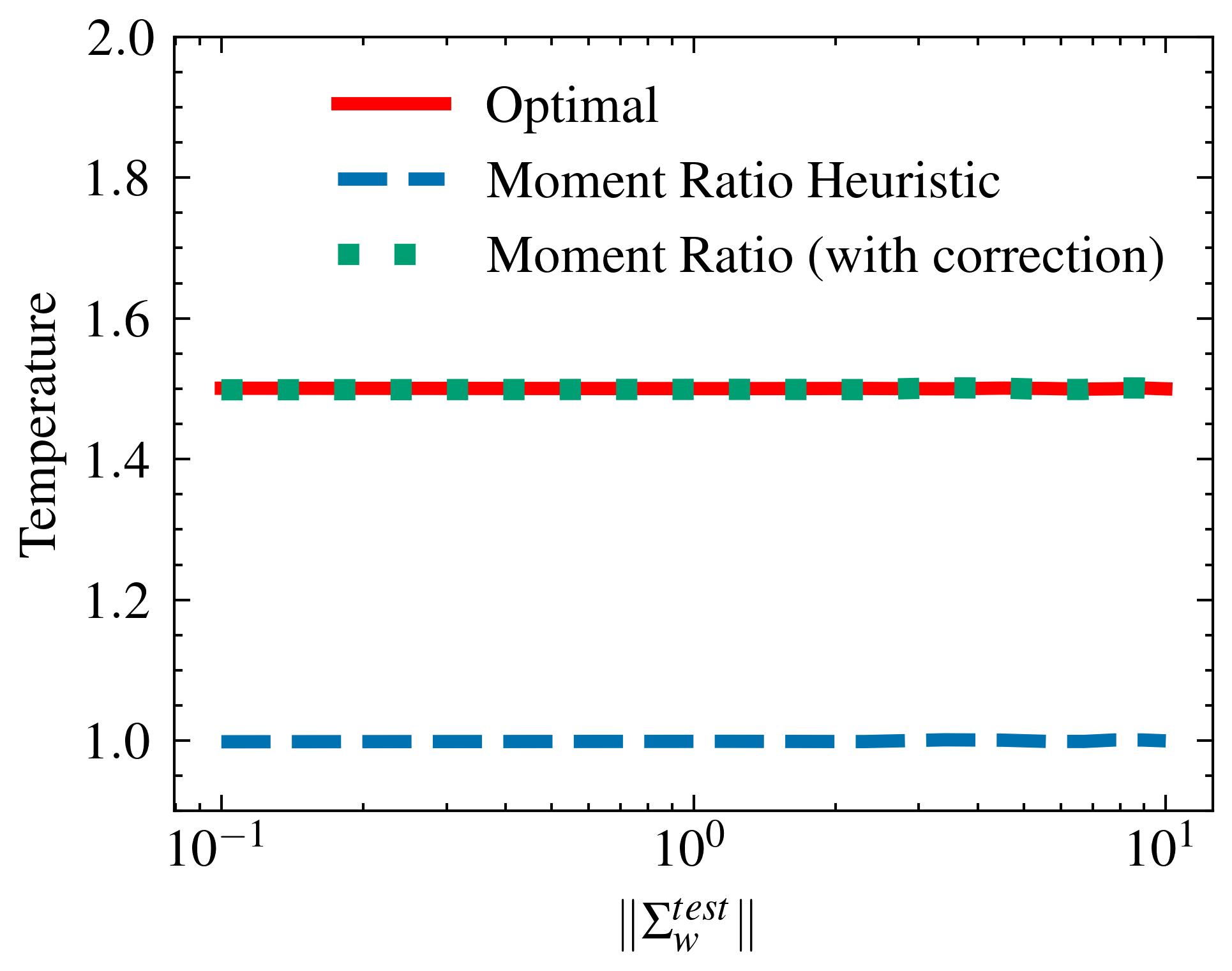}
         \captionsetup{justification=centering}
         \caption{Task Covariance Shifted\\ ($\mSigma_w^{test}= b \mSigma_w^{train})$}
     \end{subfigure}
      \begin{subfigure}[b]{0.31\textwidth}
         \centering
         \includegraphics[width=0.99\linewidth]{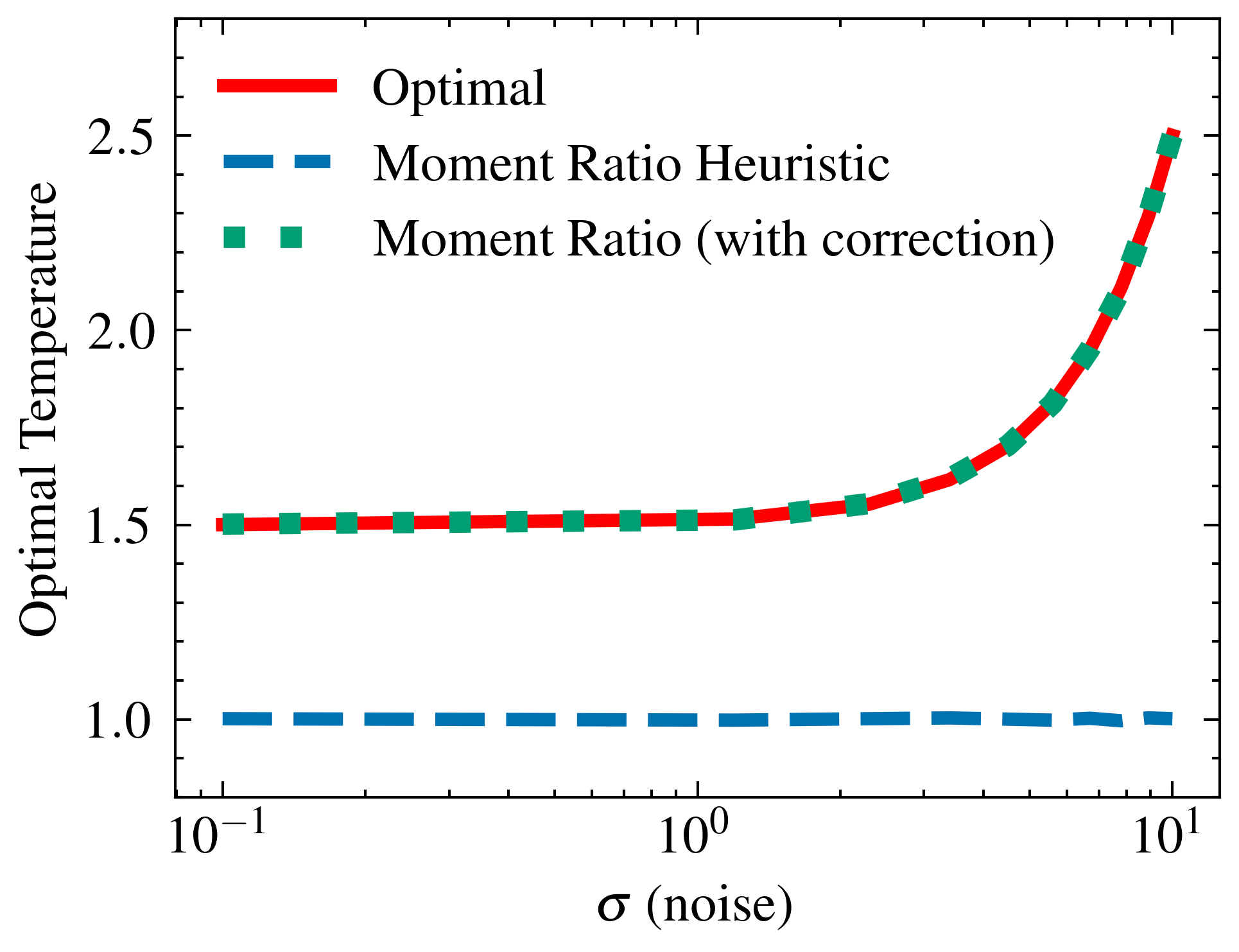}
         \captionsetup{justification=centering}
         \caption{Noise Shifted\\ ($\sigma^{test}= \sigma)$}
     \end{subfigure}
     \caption{Optimal temperature under different types of distribution shift. The moment-ratio heuristic (Appendix~\ref{appendix:insights_for_other_settings}) is derived from the closed-form optimal temperature, with its corrected form given in~\eqref{eq:moment_ratio_with_correction}. During training, we use $m = 5000$ tasks, noise level $\sigma^{train} = 0.1$, means $\vmu_x^{train} = \vmu_w^{train} = \mathbf{0}$, and covariances $\mSigma_x^{train} = \mSigma_w^{train} = \mI$. At test time, we set $\vmu_x^{test} = \vmu_w^{test} = \mathbf{0}$, $\mSigma_x^{test} = a I$, $\mSigma_w^{test} = b I$, and $\sigma^{test} = \sigma$. In each subplot, exactly one of $a$, $b$, or $\sigma$ is varied, while the other two remain fixed at their training-distribution values to isolate the effect of a single shift dimension. The dimension and context length are set to $d = 50$ and $l = 2d$.}
     \label{fig:optimal_temperature_example}
\end{figure*}

Figure~\ref{fig:optimal_temperature_example} shows the optimal temperature as a function of (i) input covariance shift, (ii) task covariance shift, and (iii) noise-variance shift. In each subplot, a single shift parameter ($a$, $b$, or $\sigma$) is varied while the others remain fixed. The closed-form optimal temperatures \eqref{eq:optimal_temperature} align closely with the moment-ratio heuristic with correction \eqref{eq:moment_ratio_with_correction}. As anticipated:
\begin{itemize}[leftmargin=*]
    \item higher input variance $a$ or noise level $\sigma$ increases the optimal temperature, while
    \item task variance $b$ does not significantly change the optimal temperature.
\end{itemize}

\begin{figure*}[h!]
    \centering
    \begin{subfigure}[b]{0.32\textwidth}
         \centering
         \includegraphics[width=0.99\linewidth]{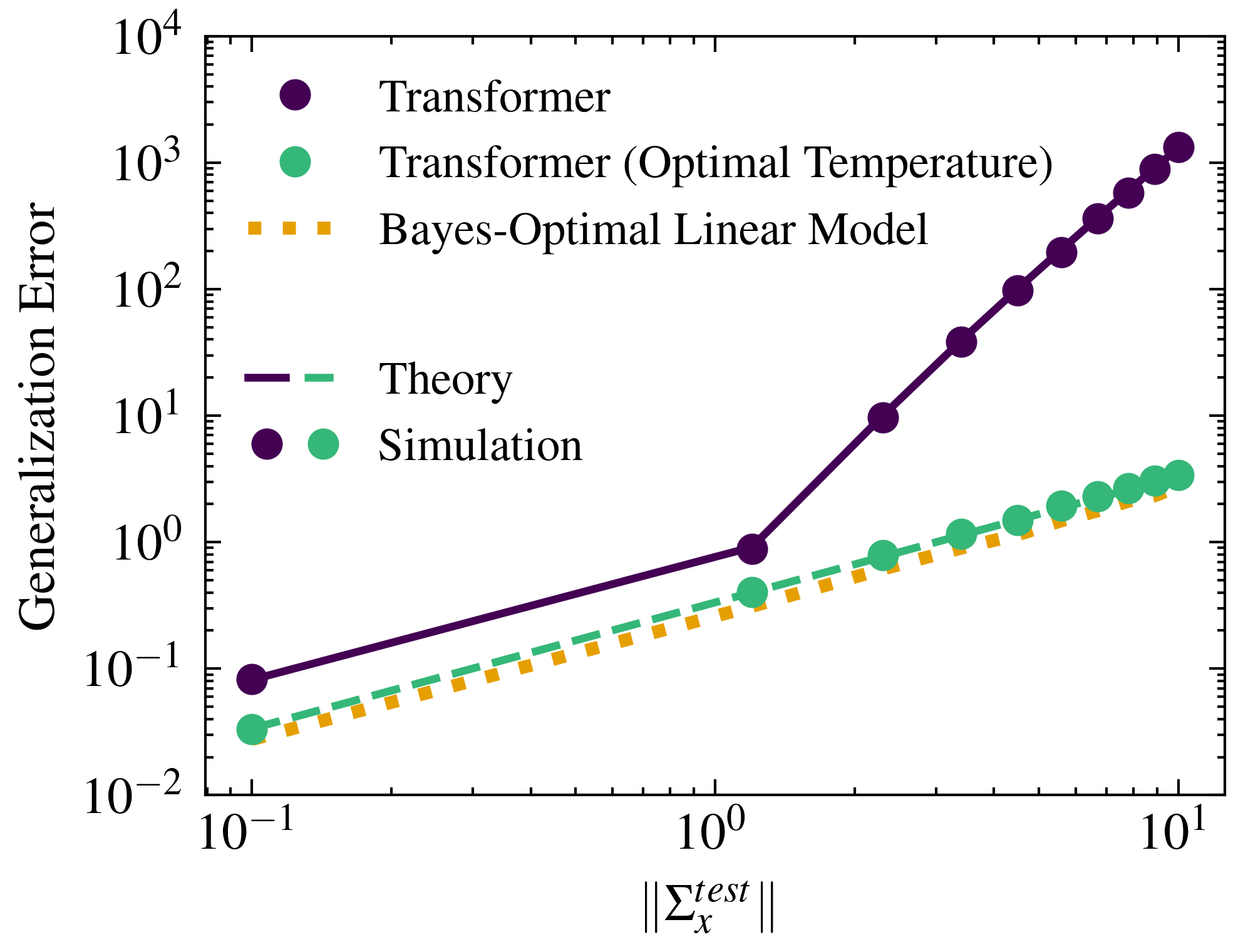}
         \captionsetup{justification=centering}
         \caption{Input Covariance Shifted\\ ($\mSigma_x^{test}= a \mSigma_x^{train})$}
     \end{subfigure}
     \begin{subfigure}[b]{0.32\textwidth}
         \centering
         \includegraphics[width=0.99\linewidth]{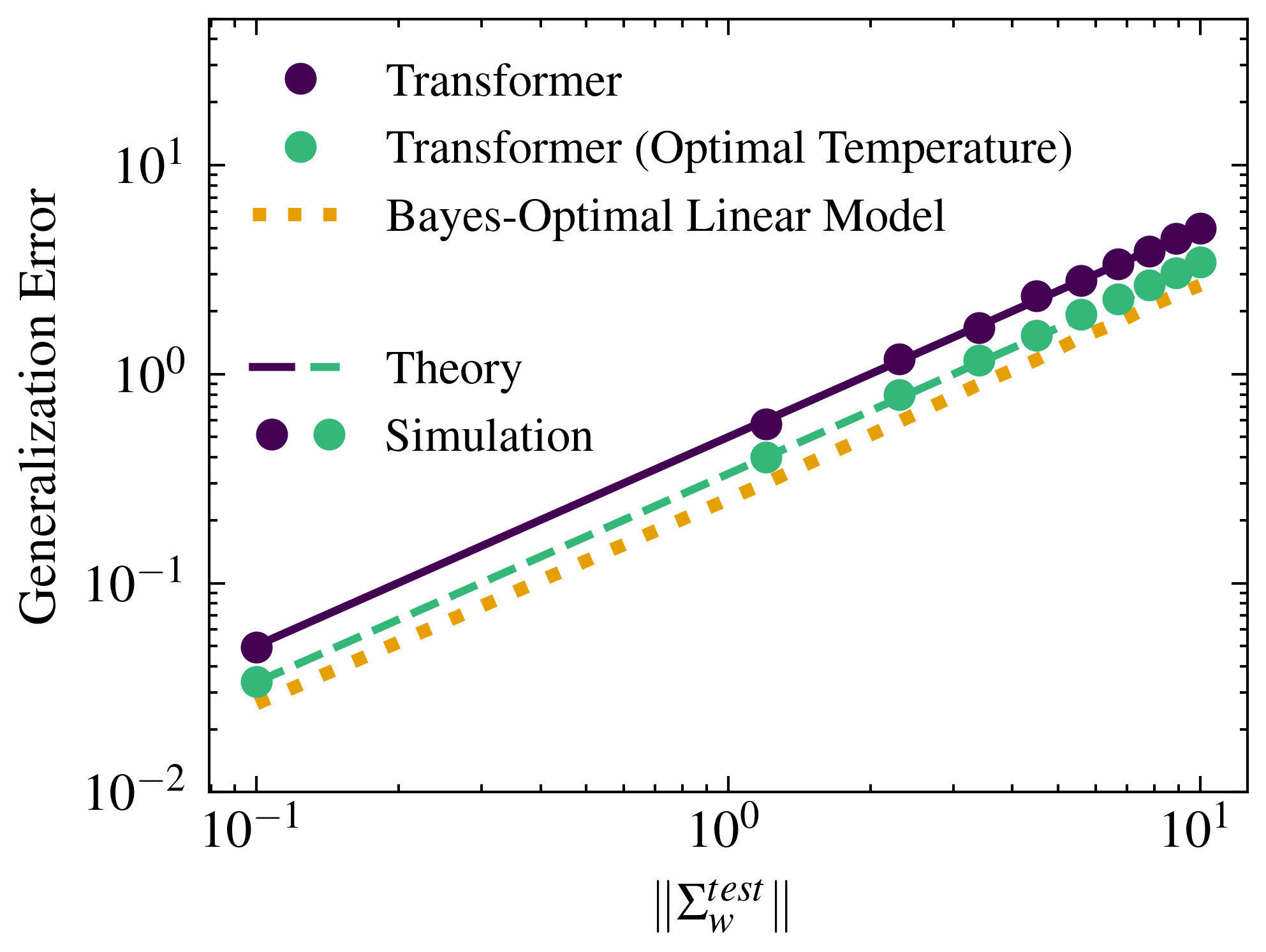}
         \captionsetup{justification=centering}
         \caption{Task Covariance Shifted\\ ($\mSigma_w^{test}= b \mSigma_w^{train})$}
     \end{subfigure}
      \begin{subfigure}[b]{0.31\textwidth}
         \centering
         \includegraphics[width=0.99\linewidth]{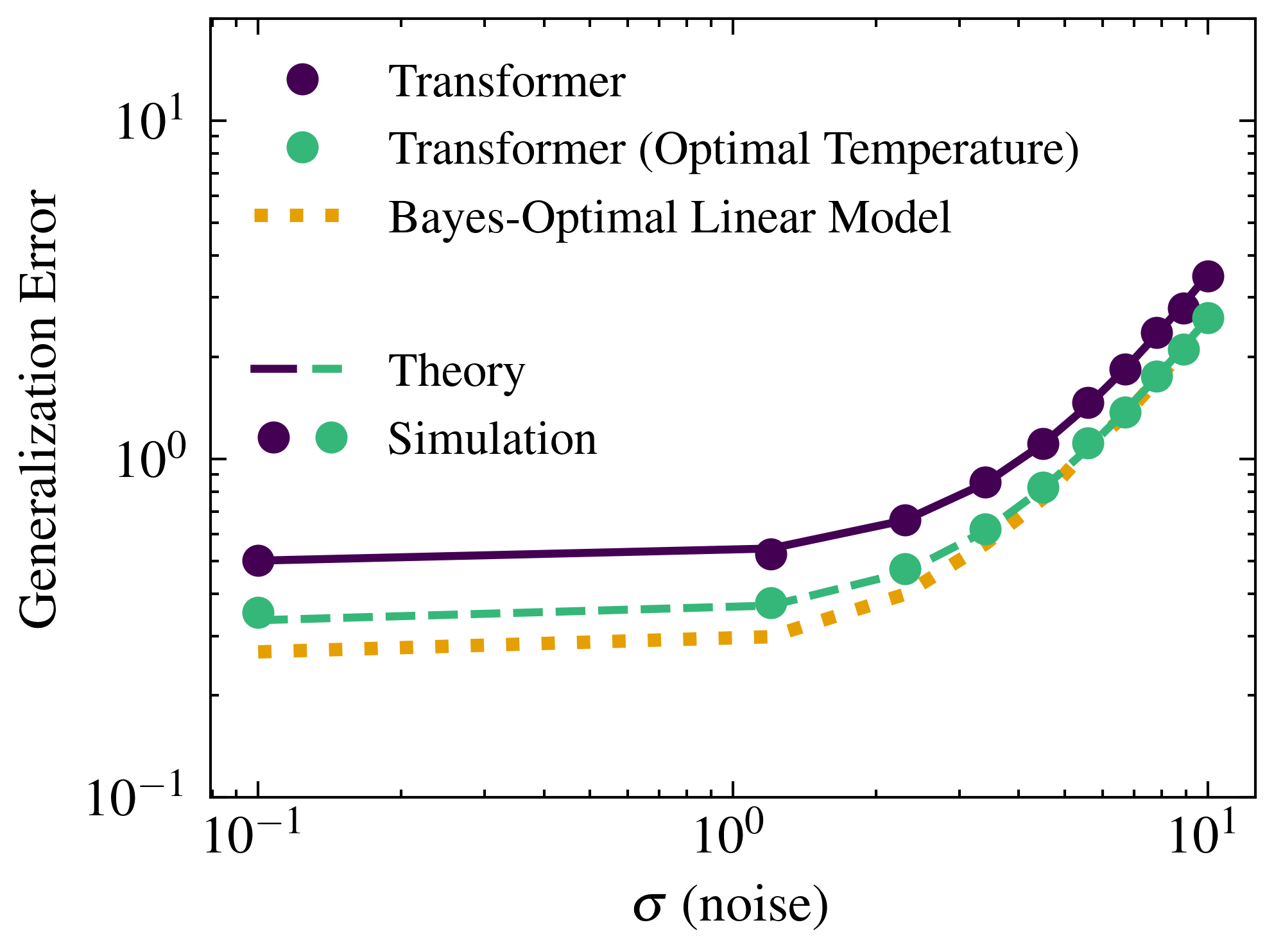}
         \captionsetup{justification=centering}
         \caption{Noise Shifted\\ ($\sigma^{test}= \sigma)$}
     \end{subfigure}
     \caption{Generalization errors corresponding to Figure \ref{fig:optimal_temperature_example}.}
     \label{fig:optimal_temperature_example_generalization}
\end{figure*}

Figure~\ref{fig:optimal_temperature_example_generalization} presents the corresponding generalization errors. These results demonstrate that the closed-form characterization accurately captures the key dependencies of the optimal temperature under a range of distribution shifts.

\end{document}